\documentclass[10pt,twocolumn,letterpaper]{article}

\usepackage[pagenumbers]{cvpr}
\usepackage{booktabs}
\usepackage{graphicx}
\usepackage{array}
\usepackage[table]{xcolor}

\definecolor{bestscore}{RGB}{248,185,185}
\definecolor{secondscore}{RGB}{255,213,165}
\definecolor{thirdscore}{RGB}{255,244,190}

\newcommand{\best}[1]{\cellcolor{bestscore}\textbf{#1}}
\newcommand{\second}[1]{\cellcolor{secondscore}#1}
\newcommand{\third}[1]{\cellcolor{thirdscore}#1}

\newcolumntype{Q}{>{\hspace{2pt}}c<{\hspace{2pt}}}

\usepackage{microtype}

\definecolor{cvprblue}{rgb}{0.21,0.49,0.74}
\usepackage[pagebackref,breaklinks,colorlinks,allcolors=cvprblue]{hyperref}
\usepackage{cuted}
\usepackage{caption}

\def\confName{CVPR}
\def\confYear{2027}

\title{IndoorMem: Floorplan as Spatial Scaffold for Consistent Indoor Video Worlds}

\author{Jia Li\textsuperscript{1} \quad Yihang Chen\textsuperscript{1,2} \quad Han Yan\textsuperscript{2} \quad Xin Ma\textsuperscript{1}\\
Yanghua Jin\textsuperscript{3} \quad Jianfei Cai\textsuperscript{1} \quad Tien-Tsin Wong\textsuperscript{1}\\[0.3em]
\textsuperscript{1}Monash University \quad \textsuperscript{2}SJTU \quad \textsuperscript{3}Spellbrush
}

\begin{document}
\maketitle

\begin{strip}
    \par
    \vspace{-45pt}
    \centering
    \includegraphics[
        width=0.98\textwidth
    ]{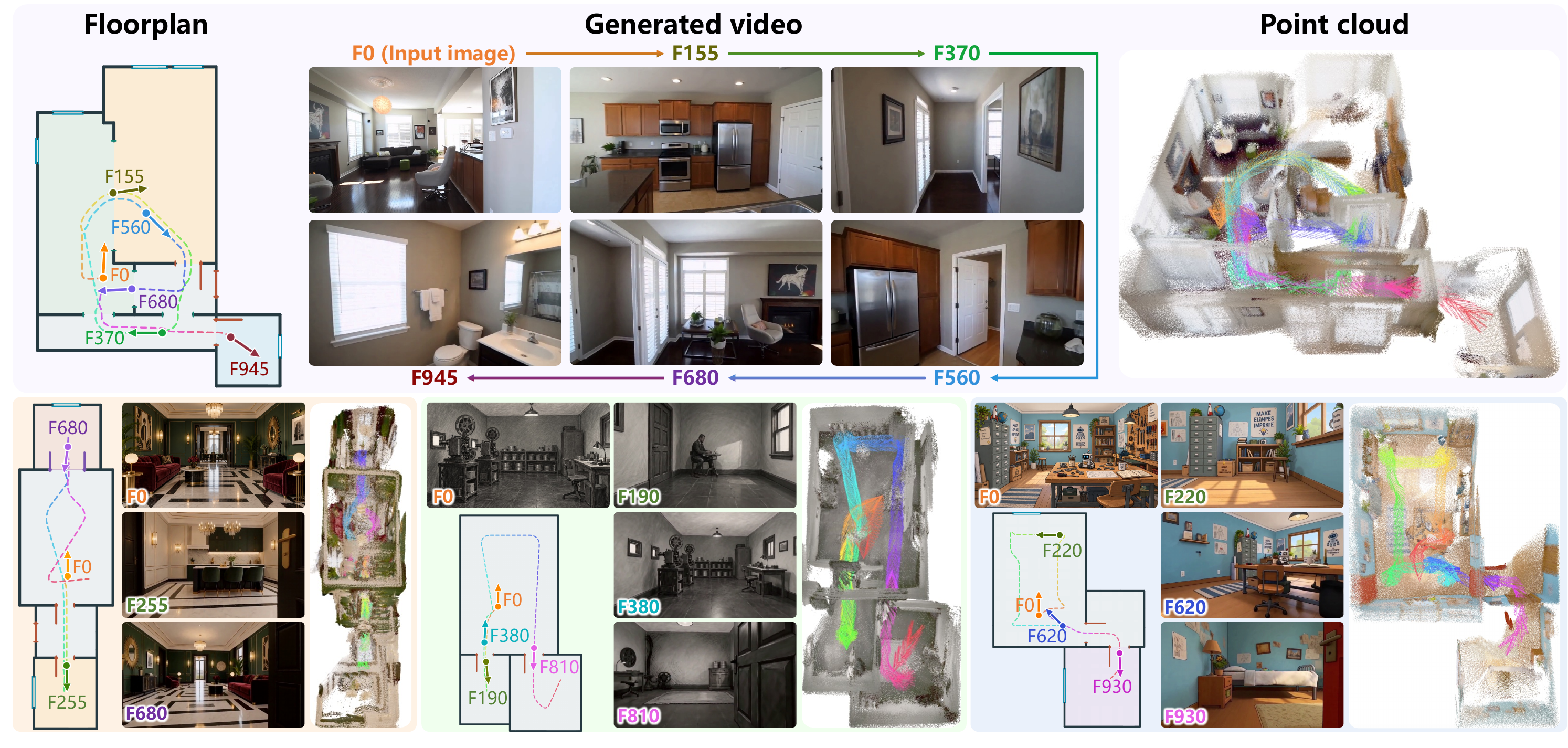}

    \captionsetup{hypcap=false}
    \captionof{figure}{Given an indoor image and exploration trajectory, \textbf{IndoorMem} generates consistent long-horizon indoor videos using a coarse floorplan as both a spatial memory index and a structural blueprint. It preserves scene appearance and structure during extensive exploration and revisits with large viewpoint changes across diverse indoor scenes, as illustrated by the generated videos and reconstructed point clouds.}
    \label{fig:teaser}
    \captionsetup{hypcap=true}
\end{strip}

\begin{abstract}
Generating immersive indoor worlds from a single image has become feasible with recent video generation models. However, maintaining scene consistency during long-horizon exploration remains challenging, especially when revisiting previously observed rooms from different viewpoints. Existing memory mechanisms either rely on potentially inaccurate 3D reconstruction or lack explicit awareness of indoor layouts and occlusions, limiting reliable spatial memory retrieval. Moreover, these methods provide limited structural guidance for unseen regions, potentially leading to structural collapse and inaccurate camera navigation. 
In this paper, we present \textbf{IndoorMem}, a unified framework that uses a coarse floorplan as a spatial scaffold for consistent indoor video generation. Our key insight is that this floorplan can serve as both a spatial index for recalling the past and a structural blueprint for guiding the future. Building upon this scaffold, we introduce Tile-indexed Memory for efficient, room-level occlusion-aware history retrieval, and Layout-aware Guidance to provide coarse geometric and semantic cues for camera control and architectural consistency. Furthermore, we propose Spatially Masked Memory Attention to suppress interference from spatially irrelevant historical content during generation. Extensive experiments demonstrate that \textbf{IndoorMem} achieves superior multi-view consistency and camera control accuracy while maintaining high video generation quality, enabling consistent long-horizon indoor exploration without dense 3D reconstruction. Project page: \url{https://riga2.github.io/indoormem}.


\end{abstract}    

\section{Introduction}
\label{sec:intro}


Imagine navigating an indoor scene starting from a single image using recent video generation models~\cite{wan2025, MiniMaxH3_2026, chen2026sana, yuan2026helios, kong2024hunyuanvideo, agarwal2026cosmos, HaCohen2024LTXVideo}. 
After exploring several rooms, you try to return to one you visited earlier, but find yourself lost as its structure and appearance have unexpectedly changed.
This highlights a fundamental challenge in long-horizon indoor video generation: maintaining a consistent world over time. While indoor scenes may {\em appear} easier to generate than outdoor environments, maintaining a consistent indoor world is in fact very challenging:
Tightly constrained architectural layouts leave much {\em smaller tolerance for structural inconsistencies}; {\em complex and frequent occlusions} make it challenging to retrieve relevant information from past observations; and transitions between rooms bring {\em rapid context changes}, e.g., furnishings, decorations, and lighting.



\begin{figure}[t]
  \centering
    \includegraphics[width=0.94\linewidth]{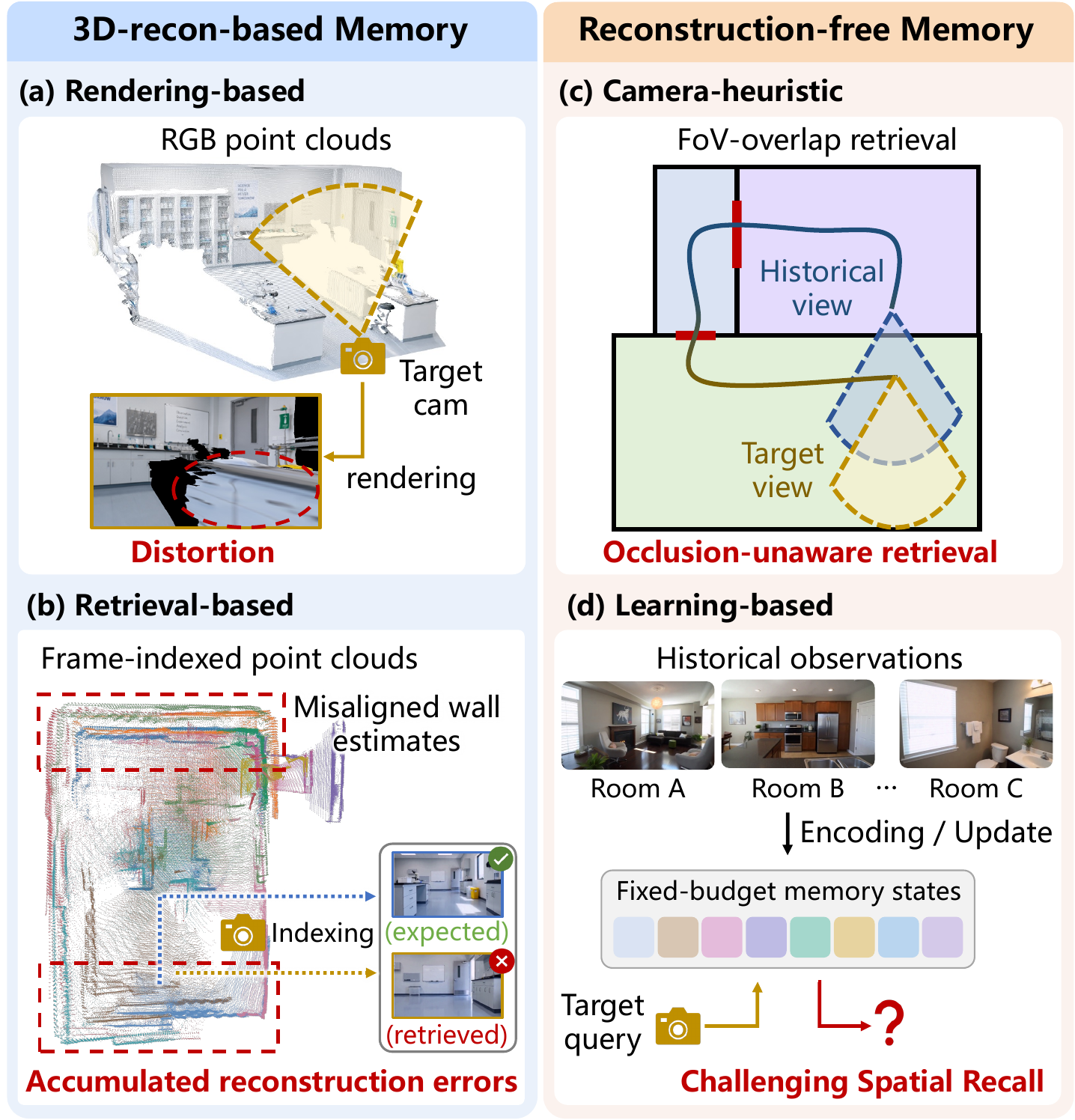}
\caption{\textbf{Limitations of representative memory mechanisms.}
Left: 3D-reconstruction-based memory relies on estimated 3D representations, which may \textbf{(a)} cause distortions during target-view rendering or \textbf{(b)} suffer from inaccurate spatial indexing due to accumulated reconstruction errors, e.g., a wall reconstructed from the yellow view incorrectly occludes the cabinet in the blue view.
Right: Reconstruction-free memory either \textbf{(c)} uses FoV overlap as a retrieval heuristic, ignoring architectural occlusions, or \textbf{(d)} encodes historical observations into memory states, making precise recall challenging without explicit spatial indexing.}
 \label{fig:memory_motivation}
\end{figure}

Existing approaches have explored various memory mechanisms to maintain long-term scene consistency, which can be broadly categorized into two groups, as illustrated in Fig.~\ref{fig:memory_motivation}. 
\textit{3D-reconstruction-based methods} incrementally construct explicit 3D representations (e.g., point clouds) from historical observations, which can either be rendered to provide visual guidance~\cite{evoke2026, yang2026neoverse, zhao2026spatia, ren2025gen3c, huang2025voyager}  or serve as spatial indices for retrieving relevant historical frames~\cite{shen2026lyra, li2025vmem, qian2026matrix, huang2025memoryforcing}. However, these approaches rely heavily on accurate geometry estimation. Inaccurate depth estimation and accumulated errors may introduce rendering artifacts (Fig.~\ref{fig:memory_motivation}(a)) or lead to inaccurate spatial indexing (Fig.~\ref{fig:memory_motivation}(b)), particularly in structurally constrained indoor environments. 
In contrast, \textit{reconstruction-free methods} either exploit camera relationships, such as field-of-view (FoV) overlap, as retrieval heuristics~\cite{team2026dreamx, sun2025worldplay, yu2025cam, xiaoworldmem, gao2026memcam}, or employ learning-based representations to encode and access historical information~\cite{wu2026infinite, yu2026memlearner, peng2026car, zhu2026sana, lingbot-world-v2, yu2026worldcrafter, li2026i3dm}. Camera-heuristic methods cannot reliably account for architectural occlusions, as overlapping viewing frustums may correspond to entirely different regions separated by walls (Fig.~\ref{fig:memory_motivation}(c)).  
Learning-based mechanisms often lack explicit spatial indexing, making it difficult to recall spatially relevant content during frequent room transitions, i.e., context changes (Fig.~\ref{fig:memory_motivation}(d)). 
In addition, generating plausible architecture in unseen regions presents another challenge. Historical observations alone do not constrain unseen layouts, risking structural collapse and inaccurate camera navigation over long sequences.



Fortunately, indoor architectural structures offer a unique opportunity:
room layouts and wall boundaries can be compactly represented by a {\em lightweight 2D floorplan}, enabling room-level visibility awareness. 
Our key insight is that a coarse floorplan can serve as {\em both a spatial index for recalling the past and a structural blueprint for guiding the future}. Motivated by this insight, we propose \textbf{IndoorMem}, a unified floorplan-based framework for consistent long-horizon indoor video generation without dense 3D reconstruction.

Specifically, starting from a single input image, we leverage an off-the-shelf multimodal large language model (MLLM)~\cite{qwen38, OpenAI2026ChatGPT} to synthesize a plausible indoor floorplan  {\em without any furniture} (like the one in Fig.~\ref{fig:teaser}). 
This preserves visible architectural structures from the input image while extending the layout into unobserved regions. 
This lightweight representation provides a persistent spatial reference for both historical memory organization and future exploration.
It constrains only the coarse architectural layout, leaving the video model free to generate furniture, materials, lighting, and fine details in unobserved regions while preserving previously generated content during revisits.

Building upon the floorplan, we introduce \textbf{Tile-indexed Memory}, which associates historical observations with spatial tiles in the discretized floorplan. Given a target trajectory, we identify visible tiles through floorplan-based visibility queries and retrieve relevant historical frames using spatial coverage-based selection. Our coarse floorplan captures room-level visibility, accounting for architectural occlusions while leaving finer-grained object-level occlusions to the video generation model. To further leverage the floorplan for future exploration, we propose \textbf{Layout-aware Guidance}, which extracts coarse ego-view geometric and semantic cues to guide camera navigation and maintain consistency with the underlying architectural layout, including in unobserved regions. Finally, to effectively incorporate retrieved memories into the generation model, we introduce \textbf{Spatially Masked Memory Attention}, which leverages coarse geometric correspondences to identify and mask irrelevant historical tokens, reducing interference from unrelated scene content. Together, these components enable the shared floorplan to support both consistent historical recall and structurally guided future exploration.

In summary, our contributions are as follows:
\begin{itemize}
    \item We introduce \textbf{IndoorMem}, a framework for consistent long-horizon indoor video generation that uses a coarse 2D floorplan as a unified spatial scaffold for recalling historical observation and guiding future exploration.

    \item We propose \textbf{Tile-indexed Memory} for efficient, room-level occlusion-aware retrieval through floorplan-based spatial indexing, and \textbf{Layout-aware Guidance} to provide coarse geometric and semantic cues for camera navigation while preserving architectural consistency.


    \item We develop \textbf{Spatially Masked Memory Attention}, which leverages coarse spatial correspondences to exclude irrelevant memory tokens for the target view, reducing interference from unrelated historical content.
\end{itemize}

\section{Related Work}
\label{sec:related}
\paragraph{Consistent Video Generation.}
Existing memory mechanisms for consistent video generation can be broadly categorized into 3D-reconstruction-based and reconstruction-free approaches (Fig.~\ref{fig:memory_motivation}). 3D-reconstruction-based methods incrementally construct explicit representations~\cite{wang2025vggt, wang2026vggt, lin2025da3} from generated observations. Among them, rendering-based methods render reconstructed scenes into target views as visual guidance~\cite{huang2025voyager, ren2025gen3c, yu2024viewcrafter, yang2026neoverse, zhao2026spatia, evoke2026}, but inaccurate geometry may introduce rendering artifacts and structural distortions~\cite{li2026i3dm, chen2025postcam} (Fig.~\ref{fig:memory_motivation}(a)). Retrieval-based methods instead use estimated geometry to spatially index historical observations and retrieve frames covering target-visible regions~\cite{shen2026lyra, li2025vmem, qian2026matrix, huang2025memoryforcing, yu2026mosaicmem}. However, reconstruction errors and scale inconsistencies may cause structural misalignments and incorrect memory retrieval (Fig.~\ref{fig:memory_motivation}(b)).

In contrast, reconstruction-free methods rely on camera heuristics or learning-based mechanisms. Camera-heuristic methods exploit FoV overlap for historical frame selection~\cite{yu2025cam, sun2025worldplay, xiaoworldmem}, but overlook architectural occlusions (Fig.~\ref{fig:memory_motivation}(c)). Learning-based methods compress historical observations into memory tokens~\cite{wu2026infinite, peng2026car}, maintain recurrent or attention-based states~\cite{zhu2026sana, lingbot-world-v2}, or adaptively query history using learned tokens~\cite{yu2026memlearner}. However, these approaches often lack explicit spatial indexing, making precise spatial recall challenging (Fig.~\ref{fig:memory_motivation}(d)). Other methods combine these two approaches~\cite{yu2026worldcrafter} or evaluate relevance over the entire history~\cite{li2026i3dm}, incurring heavy retrieval costs.

These limitations are pronounced in indoor scenes with constrained layouts, complex occlusions, and frequent context changes, motivating our floorplan-based memory.


\paragraph{Indoor Scene Generation.}
Existing indoor scene generation methods typically rely on multi-view generation and 3D reconstruction. Early methods reconstruct textured meshes from floorplans and reference images~\cite{vidanapathirana2021plan2scene}, fuse estimated depth from generated views~\cite{hollein2023text2room}, or reconstruct panoramic NeRFs~\cite{mildenhall2021nerf, fang2025ctrl}. However, these methods suffer from limited visual realism or geometric distortions.
Recently, HouseCrafter~\cite{chen2025housecrafter} and SpatialGen~\cite{fang2026spatialgen} leverage floorplans and geometric layouts for multi-view RGB-D generation and 3D reconstruction. The former produce holes during depth fusion, while the latter primarily focuses on single-room settings.
WorldMesh~\cite{schneider2026worldmesh} further employs a global mesh scaffold and 3DGS~\cite{kerbl3Dgaussians} for multi-room reconstruction, but still exhibit rendering artifacts and temporal flickering.
In contrast, our method uses a coarse floorplan for memory retrieval and structural guidance, enabling direct consistent video generation without constructing a complete 3D scene.

\section{Method}
\begin{figure*}[t]
  \centering
\includegraphics[width=1.0\textwidth]{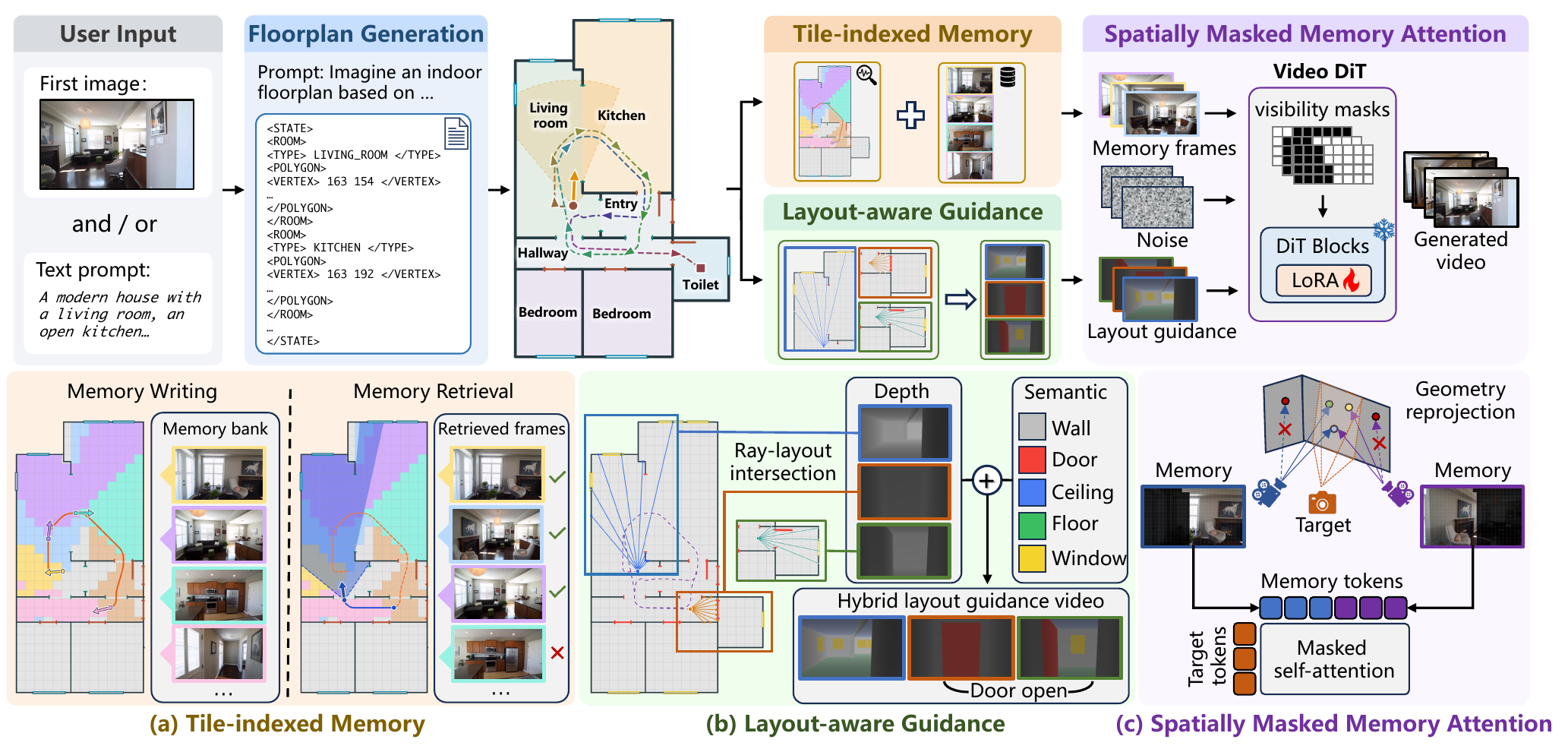}
  \caption{\textbf{Top:} \textbf{Overview of IndoorMem}. We first leverage an MLLM to synthesize a plausible floorplan from user input, which serves as a shared spatial scaffold for Tile-indexed Memory and Layout-aware Guidance. The retrieved memories and layout guidance are incorporated into a video DiT with Spatially Masked Memory Attention for consistent indoor video generation. \textbf{Bottom: Three key components.}  
  (a) \textit{Tile-indexed Memory} organizes historical observations through spatial indexing and retrieves relevant frames based on architectural visibility; 
  (b) \textit{Layout-aware Guidance} generates hybrid geometric and semantic cues via ray-layout intersection; 
  (c) \textit{Spatially Masked Memory Attention} uses coarse geometric reprojection to construct token-level visibility masks for selective memory conditioning.}
  \label{fig:overview}
\end{figure*}

We present \textbf{IndoorMem}, a floorplan-centric framework for consistent indoor video generation, as illustrated in Fig.~\ref{fig:overview}. Given an initial indoor image or text prompt, we initialize a plausible floorplan (Sec.~\ref{subsec:floorplan}) as a shared spatial scaffold. Based on this representation, \textbf{Tile-indexed Memory} (Sec.~\ref{subsec:tile}) manages historical information, while \textbf{Layout-aware Guidance} (Sec.~\ref{subsec:layout}) guides future exploration. \textbf{Spatially Masked Memory Attention} (Sec.~\ref{subsec:maksed}) enables the model to attend to spatially relevant historical content.



\subsection{Floorplan Initialization}
\label{subsec:floorplan}
To generate a plausible and consistent indoor world, we adopt a \textit{global-to-local} generation paradigm, where a lightweight 2D floorplan first establishes the global architectural layout before the video generation model synthesizes fine-grained visual details. To achieve this, we leverage an off-the-shelf multimodal large language model (MLLM)~\cite{OpenAI2026ChatGPT, qwen38} to synthesize a furniture-free floorplan from an initial indoor image or text prompt. For image inputs, the MLLM is instructed to preserve visible architectural structures while extending the layout into unobserved regions. 
Our goal is not to accurately reconstruct the entire environment, but to establish a coherent architectural layout that guides subsequent video generation. 

Specifically, we follow the Floorplan Markup Language (FML)~\cite{Shiohara_2026_CVPR}, which represents floorplans as structured token sequences describing spatial layouts and architectural elements (e.g., \texttt{<ROOM>} and \texttt{<DOOR>}). To support flexible indoor world generation, we extend the FML with open-vocabulary room types and additional attributes, including wall heights (\texttt{<HEIGHT>}), windows (\texttt{<WINDOW>}), and a reference camera configuration (\texttt{<VIEW>}) that anchors the initial viewpoint. 
Given the input and FML specification, the MLLM generates a structured token sequence that can be parsed into a floorplan (Fig.~\ref{fig:overview} (top)), serving as a shared spatial reference for subsequent Tile-indexed Memory and Layout-aware Guidance. The complete FML specification and examples are provided in the supplementary material.

\subsection{Tile-indexed Memory}
\label{subsec:tile}
After initializing the floorplan, we first use it as a persistent spatial index to organize historical observations and retrieve relevant memories for target camera trajectories. 


\paragraph{Memory Writing.}


We maintain a memory bank $\mathcal{B}=\{(I_i,P_i)\}_{i=1}^{N}$ to store historical observations, where $I_i$ and $P_i$ denote the visual content and camera parameters of frame $i$, respectively. To spatially index these observations, we discretize the floorplan into $M$ grid tiles $\mathcal{T}=\{c_j\}_{j=1}^{M}$. Each tile $c\in\mathcal{T}$ maintains an observation list $\mathcal{L}(c)$ of historical frames that observe the corresponding region (Fig.~\ref{fig:overview}(a) Left).


For newly generated frame $i$, we project each tile center onto the camera view and perform 2D ray casting on the floorplan to check frustum visibility and architectural occlusions. For each visible tile $c$, we compute an observation weight $\psi_i(c)$ to measure how well it is observed by frame $i$:
\begin{equation}
\label{eq:weight}
    \psi_i(c)=\exp\left(
    -\lambda_a a_i(c)
    -\lambda_d d_i(c)
    \right),
\end{equation}
where $a_i(c)$ and $d_i(c)$ denote the angular offset from the view center and camera-to-tile distance, respectively, with $\lambda_a$ and $\lambda_d$ controlling the decay rates, favoring observations closer to the view center and the camera.


We then add an entry $(i, \psi_i(c))$ to the observation list $\mathcal{L}(c)$. If a tile overlaps any wall boundaries, we additionally record the boundary IDs and observed sides (front/back) intersected by rays cast from frame $i$. 

\paragraph{Memory Retrieval.}
Given a target camera trajectory $\mathcal{Q}=\{P_t\}_{t=1}^{L}$, our retrieval has three stages: spatial candidate lookup, relevance scoring, and coverage-based selection.



First, we apply the same visibility test to the target trajectory to obtain the visible tiles $\mathcal{T}_{\mathcal{Q}}$. By querying the observation list of each target-visible tile, we obtain the candidate historical frame indices:
$\mathcal{C}_{\mathcal{Q}}=
\bigcup_{c\in\mathcal{T}_{\mathcal{Q}}}
\operatorname{Index}(\mathcal{L}(c))$.
This tile-indexed lookup restricts subsequent relevance scoring to candidate frames, avoiding evaluating the entire history.

Next, we evaluate how well each candidate frame observes the target-visible regions by considering both historical observation quality and viewing compatibility. Specifically, for each historical candidate frame $i\in\mathcal{C}_{\mathcal{Q}}$ and target-visible tile $c$, we define its relevance as
\begin{equation}
\label{eq:relevance}
\begin{aligned}
    r_i(c\mid\mathcal{Q})
    &=\psi_i(c)\,\kappa_i(c\mid\mathcal{Q}),\\
    \kappa_i(c\mid\mathcal{Q})
    &=\sqrt{
    w_{\mathrm{pos}}(\Delta{d}_i(c))
    \cdot
    w_{\mathrm{angle}}(\Delta\theta_i(c))
    },
\end{aligned}
\end{equation}
where $\psi_i(c)$ is the stored observation weight (Eq.~\ref{eq:weight}), and $\kappa_i(c\mid\mathcal{Q})$ measures viewing compatibility between historical frame $i$ and the target camera in $Q$ that observes the same tile $c$. Here, $\Delta d_i(c)$ and $\Delta\theta_i(c)$ denote the camera distance and viewing angular difference relative to the representative target camera observing tile $c$, selected based on normalized target observation weights along $\mathcal{Q}$. $w_{\mathrm{pos}}$ and $w_{\mathrm{angle}}$ are bounded exponential decay functions, favoring historical observations captured from nearby and similarly oriented viewpoints. For tiles overlapping wall boundaries, we also enforce matching boundary IDs and observed sides to ensure the same observed wall surfaces.

Finally, we employ a greedy maximum-coverage selection under a memory budget of $K$. Starting with $\mathcal{S}=\{i_0\}$, where $i_0$ is the initial frame index, we define the current coverage $H_{\mathcal{S}}(c)$ of each target-visible tile $c$ and the marginal gain $\Delta(j\mid\mathcal{S})$ of adding a candidate frame $j$ as:
\begin{equation}
\label{eq:max_cov}
\begin{aligned}
H_{\mathcal{S}}(c)
&= \max_{i\in\mathcal{S}} r_i(c\mid\mathcal{Q}),\\
\Delta(j\mid\mathcal{S})
&= \sum_{c\in\mathcal{T}_{\mathcal{Q}}}
\max\left(
r_j(c\mid\mathcal{Q})-H_{\mathcal{S}}(c),\,0
\right).
\end{aligned}
\end{equation}
We iteratively select the candidate with the largest marginal gain and update $\mathcal{S}$ accordingly, until reaching the memory budget or a stopping threshold. The selected frames are then retrieved from the memory bank $\mathcal{B}$ to condition subsequent video generation. Details are provided in the appendix.

\subsection{Layout-aware Guidance}
\label{subsec:layout}
While the memory mechanism preserves historical observations for consistent revisiting, it cannot guide camera exploration. 
Existing camera-controlled video generation methods typically use Plücker-ray embeddings~\cite{he2024cameractrl, bahmani2025ac3d} or camera-aware positional encodings~\cite{zhang2026unified, yin2026scope}, providing limited explicit structural guidance. 
Instead, we leverage the initialized floorplan and translate user-specified camera trajectories into coarse ego-view geometric and semantic guidance, enabling accurate camera control while maintaining architectural consistency during future exploration.


\paragraph{Coarse Geometric Guidance.}
To extract ego-view geometric guidance, we first construct a coarse 3D architectural proxy from the 2D floorplan. Taking the world $Y$-axis as the vertical axis, we extrude architectural boundaries into vertical wall surfaces with height $H$, and add floor and ceiling surfaces at $Y=0$ and $Y=H$, respectively.

Given the target camera intrinsics and pose, we cast a world-space ray $\mathbf{x}_p(t)=\mathbf{o}+t\mathbf{d}_p$ for each image pixel $p$, where $\mathbf{o}$ and $\mathbf{d}_p$ denote the camera origin and ray direction, respectively. We compute its intersections with the wall, floor, and ceiling surfaces using ray-plane intersection tests.
Let $\mathcal{Z}_p$ denote the set of valid intersection depths. We construct a normalized depth map as coarse geometric guidance by selecting and normalizing the nearest valid intersection: $D(p)=\operatorname{Norm}(\min\mathcal{Z}_p)\in[0,1]$, as shown in Fig.~\ref{fig:overview}(b).


\paragraph{Hybrid Layout Guidance.}
While the depth sequence conveys coarse architectural structure and camera motion, multi-room exploration also requires architectural semantics for door interactions and room transitions. We therefore augment the geometric guidance with five architectural semantic labels: wall, door, floor, ceiling, and window. 

During ray-surface intersection, each pixel inherits the semantic label of its nearest intersected surface, producing a semantic map aligned with the rendered depth. To incorporate both geometric and semantic information without increasing the number of input channels, we blend the normalized depth with modulated semantic colors to form hybrid guidance (shown in Fig.~\ref{fig:overview}(b)):
\begin{equation}
    \mathbf{G}(p)
    = (1-\alpha_p)\,D(p)\mathbf{e}
    + \alpha_p\,\phi(D(p))\mathbf{c}_p,
\end{equation}
where $\mathbf{c}_p$ and $\alpha_p$ denote the semantic color and blending weight, respectively, $\mathbf{e}$ is an all-ones vector, and $\phi$ is a linear brightness modulation function.

To support room transitions, we further simulate door-opening motion as the camera approaches. By rendering the animated door geometry at each frame, the resulting guidance captures the opening motion and the gradual reveal of the new room, promoting coherent room transitions.

\subsection{Spatially Masked Memory Attention}
\label{subsec:maksed}
Although Sec.~\ref{subsec:tile} retrieves relevant historical frames, not all regions within these frames are visible to the target cameras. Directly attending to such irrelevant regions may introduce content interference or unexpected repetition. To address this, we leverage the coarse 3D architectural proxy from Sec.~\ref{subsec:layout} to construct visibility masks, restricting memory attention to regions co-visible with the target cameras.

\paragraph{Geometry-guided Visibility Mask.}
For each retrieved memory frame $I_i$ with camera $P_i$, we divide the image into spatial patches matching the DiT latent. For each memory patch $m$, we sample a set of pixel coordinates $\mathcal{U}_m$ and cast rays to obtain their nearest intersections $\mathbf{x}_i(u)$ with the coarse architectural geometry, following Sec.~\ref{subsec:layout}. Each valid surface point is then reprojected to the target view $P_t$. We check whether the projection falls within the target image and whether its surface ID, observed side, and depth are consistent with the target-view rendering. We define a binary visibility indicator $\nu_{i,t}(u)\in\{0,1\}$, assigning 1 to co-visible samples, as illustrated in Fig.~\ref{fig:overview}(c). 

Due to temporal compression in video DiT, each latent $\ell$ corresponds to multiple video frames $\mathcal{J}_{\ell}$. We aggregate visibility across these frames to construct a patch-level mask:
\begin{equation}
\begin{aligned}
M_{\ell,i}(m)
&=
\mathbf{1}\left[
f_{\ell,i}(m)\geq\tau_{\mathrm{vis}}
\right],\\
f_{\ell,i}(m)
&=
\frac{1}{|\mathcal{J}_{\ell}|\,|\mathcal{U}_m|}
\sum_{t\in\mathcal{J}_{\ell}}
\sum_{u\in\mathcal{U}_m}
\nu_{i,t}(u),
\end{aligned}
\end{equation}
where $\tau_{\mathrm{vis}}$ is the visibility threshold.

\begin{table*}[t]
\centering
\caption{\textbf{Quantitative comparison}. Rev./Adj. MEt3R denote revisit/adjacent-frame MEt3R, and ZOR denotes zero-overlap rate. Overall averages SC, BC, TF, MS, and AQ.
Best, second-best, and third-best results are highlighted in red, orange, and yellow, respectively.}
\label{tab:quantitative}
\vspace{-2mm}

\setlength{\tabcolsep}{3.3pt}
\renewcommand{\arraystretch}{0.82}
\setlength{\aboverulesep}{0.2ex}
\setlength{\belowrulesep}{0.2ex}

\footnotesize

\begin{tabular}{@{}l*{6}{c}*{6}{Q}}

\toprule
& \multicolumn{3}{c}{Multi-view Consistency}
& \multicolumn{3}{c}{Camera Control}
& \multicolumn{6}{c}{Video Generation Quality} \\
\cmidrule(lr){2-4}
\cmidrule(lr){5-7}
\cmidrule(l){8-13}

Method
& Rev.\ MEt3R$\downarrow$
& ZOR (\%)$\downarrow$
& Adj.\ MEt3R$\downarrow$
& RotErr$\downarrow$
& TransErr$\downarrow$
& CamMC$\downarrow$
& SC$\uparrow$
& BC$\uparrow$
& TF$\uparrow$
& MS$\uparrow$
& AQ$\uparrow$
& Overall$\uparrow$ \\
\midrule

DreamX-World
& 0.797 & 37.35 & 0.050
& 117.93 & 2.776 & 3.699
& 75.43 & 84.80 & 95.88 & 98.37 & 45.41 & 79.98 \\

HY-WorldPlay
& \second{0.656} & \second{16.74} & 0.055
& 117.50 & 2.530 & 3.486
& 74.29 & 84.72 & \third{95.91} & \third{98.90}
& 42.63 & 79.29 \\

Alaya-EVOKE
& 0.783 & 42.25 & \third{0.047}
& 103.21 & 2.625 & 3.429
& \second{77.53} & 85.27 & 95.28 & 98.32
& \best{52.80} & \second{81.84} \\

Infinite-World
& \second{0.656} & \third{24.26} & \second{0.040}
& 127.34 & 2.596 & 3.598
& 76.44 & \third{85.58} & \best{97.40} & \best{99.23}
& 45.73 & \third{80.88} \\

Lyra 2.0
& 0.744 & 38.53 & 0.054
& \third{80.08} & 2.372 & 2.989
& 74.45 & 84.74 & 95.23 & 98.60
& 43.19 & 79.24 \\

LingBot-World-v2
& 0.870 & 52.91 & 0.093
& 123.26 & 2.941 & 3.875
& 71.11 & 82.46 & 91.42 & 95.16
& \second{49.07} & 77.84 \\

Matrix-Game-3.5
& 0.711 & 34.08 & 0.053
& \second{54.16} & \second{2.021} & \second{2.420}
& 75.40 & 83.99 & \second{96.07} & 98.64
& 45.12 & 79.84 \\

SANA-WM
& 0.824 & 46.64 & 0.064
& 105.90 & 2.710 & 3.508
& 74.96 & 84.21 & 94.95 & 98.33
& 45.92 & 79.67 \\

WorldCrafter
& \third{0.681} & 30.10 & 0.050
& 80.74 & \third{2.318} & \third{2.926}
& \third{76.57} & \second{85.73} & 95.84 & 98.74
& 45.50 & 80.48 \\


\midrule
\textbf{Ours}
& \best{0.425} & \best{5.81} & \best{0.038}
& \best{28.58} & \best{1.207} & \best{1.410}
& \best{78.61} & \best{87.08} & 95.88
& \second{99.02} & \third{48.71} & \best{81.86} \\

\bottomrule
\end{tabular}
\end{table*}


\paragraph{Masked Memory Conditioning.}
We concatenate retrieved memory tokens, layout guidance tokens, and noisy target video tokens into a sequence processed by the video DiT. 
During self-attention, we add a bias of $-\infty$ to target-to-memory attention logits where $M_{\ell,i}(m)=0$, leaving all other attention connections unchanged. Our visibility mask provides a coarse spatial constraint without exhaustively computing token-to-token correspondences, allowing the fine-tuned DiT to adaptively attend to relevant historical content while reducing interference from unrelated regions.

\begin{figure*}[t]
  \centering
\includegraphics[width=1.0\textwidth]{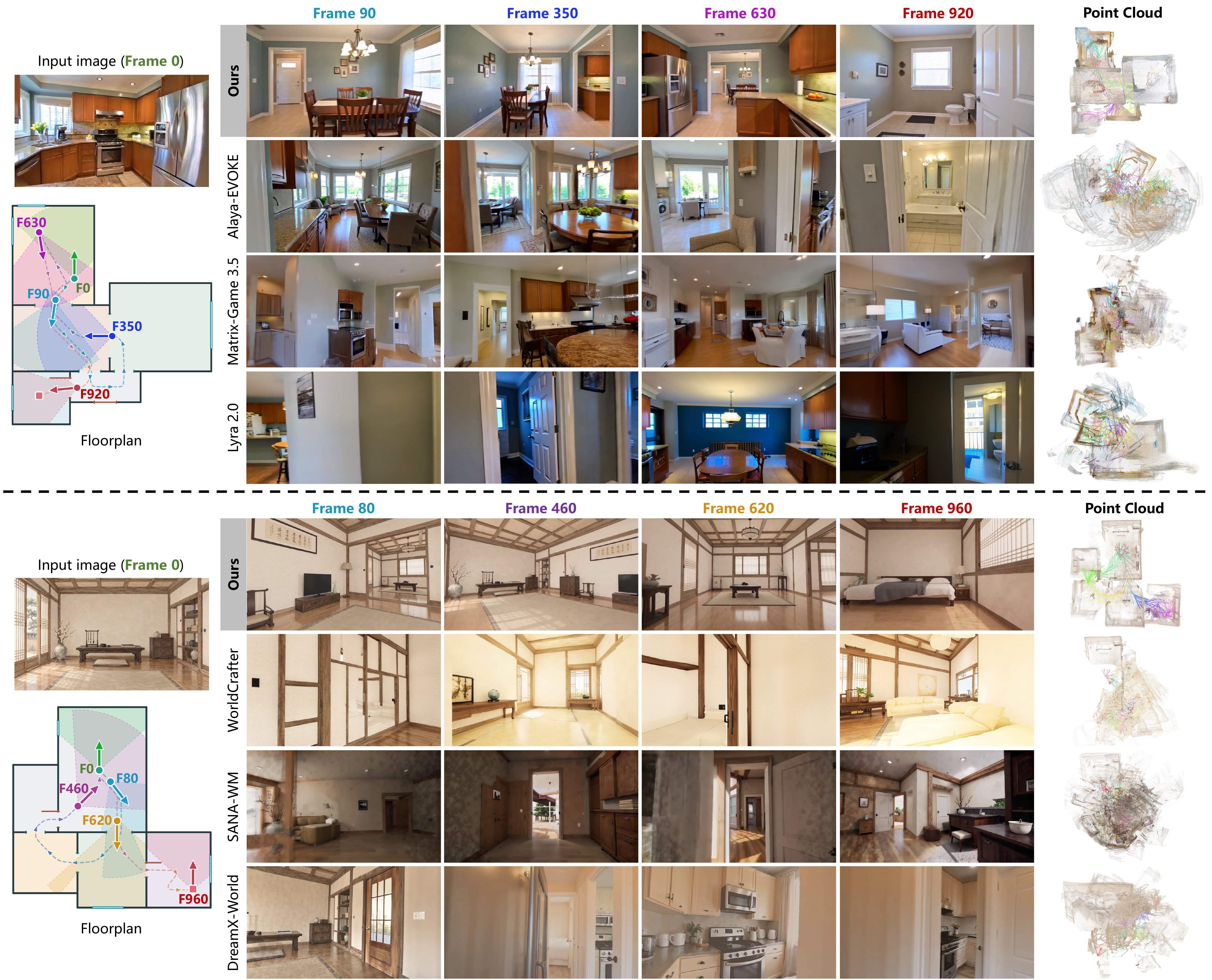}
  \caption{\textbf{Qualitative comparison} on a real-world scene (top) and a generated scene (bottom). Left: The input image (Frame 0) and the floorplan with the camera exploration trajectory. Middle: Generated videos from different methods. Right: Point clouds reconstructed from the generated videos. The colored frame indices correspond to the camera viewpoints marked on the floorplans.}
 \label{fig:comp_main}
\end{figure*}

\section{Experiments}
In this section, we first describe the experimental setup (Sec.~\ref{sec:setup}), followed by comparisons with state-of-the-art video world models (Sec.~\ref{sec:comp}) and ablation studies (Sec.~\ref{sec:ablation}). Additional details and results are provided in the appendix.

\subsection{Experimental Setup}
\label{sec:setup}
\paragraph{Datasets.}
For training, we construct a synthetic indoor exploration dataset by rendering 675 scenes from ProcTHOR~\cite{procthor}, yielding 9,516 training clips of 243 frames each, with retrieved memory frames using our memory mechanism.
For evaluation, we collect 407 initial indoor images, including 207 generated by GPT-Image 2 and 200 real images from RealEstate10K~\cite{Zhou2018} and DL3DV~\cite{ling2024dl3dv}, covering diverse indoor environments and visual styles. For each image, we employ GPT-6~\cite{OpenAI2026ChatGPT} to initialize a floorplan and construct a corresponding 972-frame camera trajectory with revisits, yielding 407 long-horizon evaluation sequences.

\paragraph{Implementation Details.}
We fine-tune the MiniMax-H3 Ref2VA~\cite{MiniMaxH3_2026} backbone using rank-32 LoRA~\cite{hu2021lora} applied to the attention and feed-forward layers of the DiT. Training runs for 1,000 steps using 243-frame clips at $832\times480$, with a learning rate of $10^{-4}$ and a batch size of eight. 
At inference, we generate videos autoregressively in 243-frame clips with one overlapping frame. We retrieve up to eight historical frames ($K=8$) using our memory mechanism. Further training and inference details are in the appendix.

\paragraph{Evaluation Metrics.} We assess performance across three dimensions: \textbf{(1) Multi-view Consistency.} 
Previous methods~\cite{li2025vmem, sun2025worldplay} typically evaluate revisit consistency using forward-and-backward camera trajectories with identical camera poses. However, such protocols may favor direct reuse of historical frames rather than testing consistency across different viewpoints, as commonly encountered during real-world exploration.
We therefore adopt MEt3R~\cite{asim2025met3r} to evaluate long-term multi-view consistency. 
Specifically, using camera trajectories and coarse 3D geometry derived from the floorplans, we precompute 40,256 pairs of views observing overlapping regions, following an ``observe-leave-revisit'' pattern. Since severe inconsistencies may prevent the geometric foundation model~\cite{cabon2025must3r} used by MEt3R from establishing valid correspondences, we additionally report the zero-overlap rate (ZOR), defined as the proportion of revisit pairs with no valid correspondences. We also report adjacent-frame MEt3R to measure short-term temporal consistency.
\textbf{(2) Camera Control Accuracy.} 
We estimate camera trajectories using VGGT-$\Omega$~\cite{wang2026vggt} and evaluate rotation error (RotErr), translation error (TransErr), and camera pose matrix discrepancy (CamMC), following SANA-WM~\cite{zhu2026sana}.
\textbf{(3) Video Generation Quality.} 
We adopt five VBench~\cite{huang2024vbench} metrics: Subject Consistency (SC), Background Consistency (BC), Temporal Flickering (TF), Motion Smoothness (MS), and Aesthetic Quality (AQ). We also report average score.

\subsection{Comparisons}
\label{sec:comp}
\paragraph{Comparison Methods.}
We compare with nine recent memory-enabled video world models, including (1) 3D-reconstruction-based methods: Alaya-EVOKE~\cite{evoke2026}, Lyra 2.0~\cite{shen2026lyra}, and Matrix-Game-3.5~\cite{qian2026matrix}; and (2) reconstruction-free methods: DreamX-World~\cite{team2026dreamx}, HY-WorldPlay~\cite{sun2025worldplay}, LingBot-World-v2~\cite{lingbot-world-v2}, Infinite-World~\cite{wu2026infinite}, SANA-WM~\cite{zhu2026sana}, and WorldCrafter~\cite{yu2026worldcrafter}. We also compare with WorldMesh~\cite{schneider2026worldmesh} and Code-World-Model~\cite{chen2026code} in the appendix.
We generate videos using their official implementations.

\paragraph{Comparison Results.} As shown in Tab.~\ref{tab:quantitative}, our method achieves the best performance in multi-view consistency, camera control accuracy, and overall generation quality. In particular, the substantial improvements in consistency and camera control demonstrate the effectiveness of our proposed memory mechanism and layout guidance strategy. 
As illustrated in Fig.~\ref{fig:comp_main}, most comparison methods struggle to maintain scene consistency during large-range exploration, especially when revisiting previously observed regions from different viewpoints. In contrast, our method preserves scene consistency even under substantial viewpoint changes, e.g., approximately $90^\circ$ (Frames 90 and 350) or $180^\circ$ (Frames 0 and 630) in Fig.~\ref{fig:comp_main} (top), and revisiting the same room through a different entrance after exploring other rooms in Fig.~\ref{fig:comp_main} (bottom) (Frames 0, 80 and 460).
The coherent point clouds reconstructed from our generated videos (Fig.~\ref{fig:comp_main}, rightmost column) and well-aligned camera poses further demonstrate our ability to maintain consistent scene structures and accurate camera control in long-horizon exploration.


\begin{table}[tp]
\centering
\begin{minipage}{\linewidth}

\centering
\caption{\textbf{Ablation study} of the memory mechanism (top) and layout-aware guidance (bottom). }
\label{tab:ablation}
\vspace{-2mm}

\setlength{\tabcolsep}{1.2pt}
\renewcommand{\arraystretch}{0.82}
\setlength{\aboverulesep}{0.2ex}
\setlength{\belowrulesep}{0.2ex}
\small

\resizebox{\columnwidth}{!}{%
\begin{tabular}{@{}l*{5}{c}@{}}
\toprule
& \multicolumn{2}{c}{Multi-view Consistency}
& \multicolumn{3}{c}{Camera Control} \\
\cmidrule(lr){2-3}
\cmidrule(l){4-6}

Method
& Rev.\ MEt3R$\downarrow$
& ZOR (\%)$\downarrow$
& RotErr$\downarrow$
& TransErr$\downarrow$
& CamMC$\downarrow$ \\
\midrule

w/o Memory
& 0.715 & 38.38
& 64.58 & 1.932 & 2.436 \\

w/o Masked Attn.
& 0.449 & 8.62
& 41.80 & 1.492 & 1.802 \\

FoV-based Retri.
& \second{0.415} & \second{5.29}
& \third{36.56} & \third{1.287} & \third{1.567} \\

Geo-based Retri.
& 0.446 & 6.77
& {38.76} & {1.326} & {1.625} \\

I3DM-based Retri.
& \third{0.423} & \third{5.62}
& \second{31.11} & \second{1.225} & \second{1.457} \\

\midrule

w/o Layout Guid.
& 0.505 & 15.30
& 96.82 & 2.204 & 2.995 \\

w/o Semantic Label
& 0.443 & 8.30
& 45.31 & 1.457 & 1.811 \\

\midrule

\textbf{Ours}
& \best{0.384} & \best{2.24}
& \best{22.54} & \best{1.009} & \best{1.177} \\

\bottomrule
\end{tabular}%
}
%
\end{minipage}
\par\vspace{3pt}
\begin{minipage}{\linewidth}
\captionsetup{type=figure}
\centering
\includegraphics[width=1.0\linewidth]{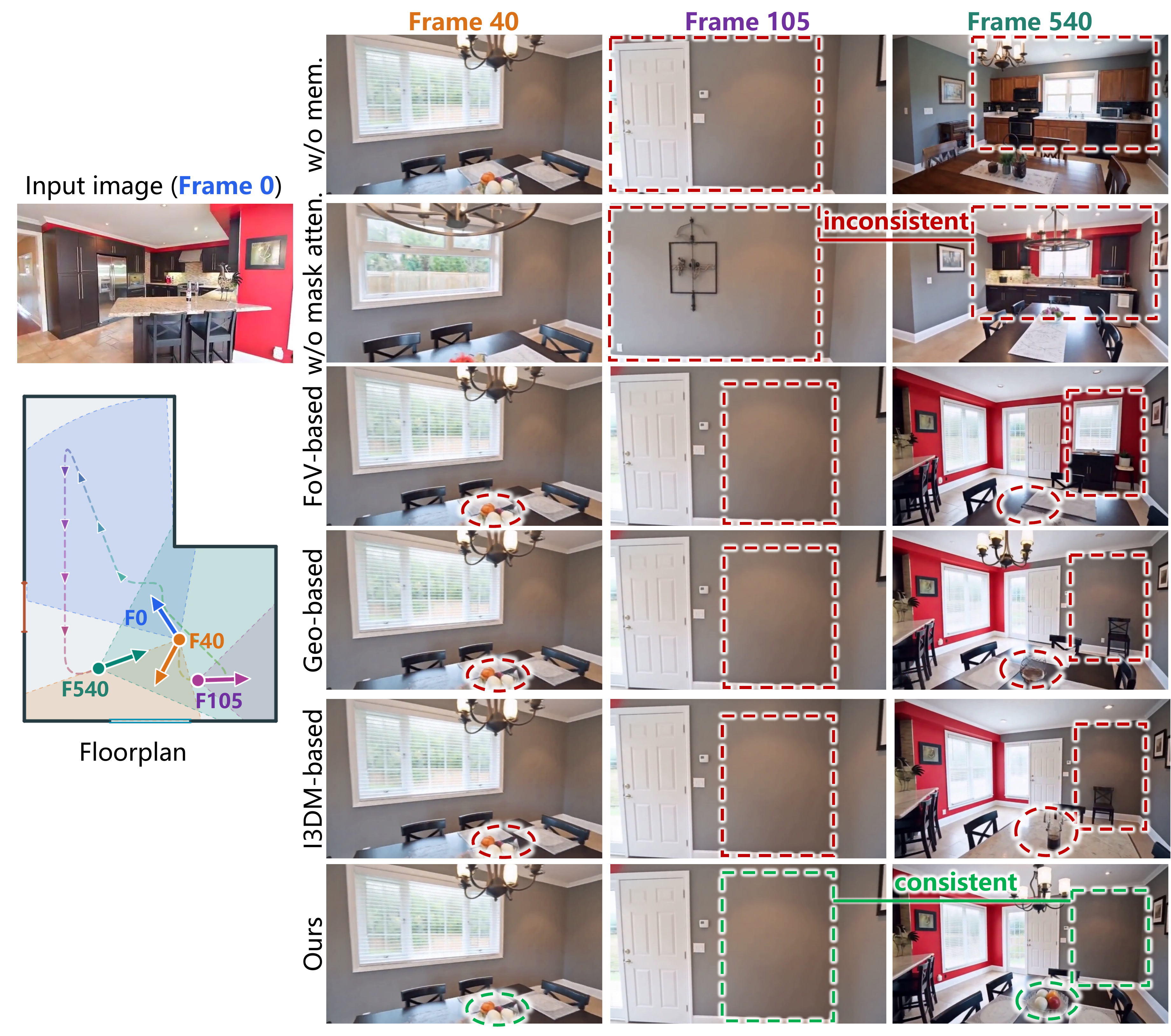}
  \caption{
  \textbf{Ablation study of memory mechanism.}     
Red dashed boxes and circles highlight inconsistent regions, while green ones indicate consistent results from our full method.
  }
 \label{fig:ab_retrieval}
\end{minipage}
\par\vspace{3pt}
\begin{minipage}{\linewidth}
\captionsetup{type=figure}
\centering
\includegraphics[width=1.0\linewidth]{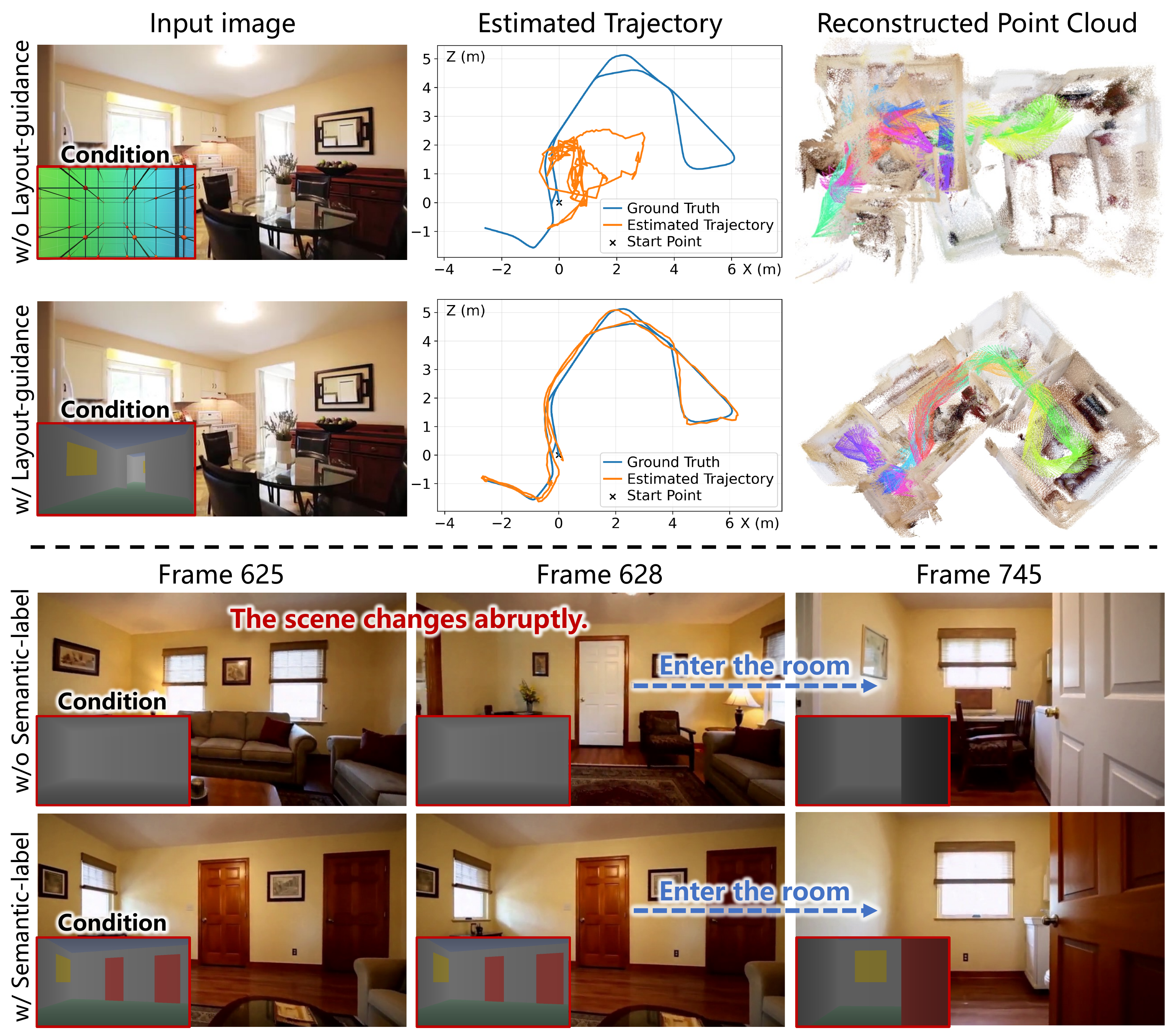}
  \caption{
  \textbf{Ablation study of layout-aware guidance.} 
  Top: Layout-aware guidance improves camera control and architectural consistency, yielding aligned trajectories and coherent point clouds.
  Bottom: Without semantic guidance, the model abruptly hallucinates a door on existing scene during room transitions.
  }
 \label{fig:ab_guidance}
\end{minipage}
\end{table}

\subsection{Ablation Study}
\label{sec:ablation}


We sample 107 indoor scenes from our test set, including both real-world and generated scenes, for the ablation study. Each variant modifies only one component while keeping the remaining modules and experimental settings fixed.

\paragraph{Memory Mechanism.}
We construct five ablation variants:
(1) \textit{w/o Memory}, which uses only the initial frame or the last frame of the preceding clip as condition;
(2) \textit{w/o Masked Attn.}, which replaces our Spatially Masked Memory Attention with standard full attention;
(3) \textit{FoV-based Retrieval}, which adopts FoV-overlap-based retrieval following WorldMem~\cite{xiaoworldmem};
(4) \textit{Geo-based Retrieval}, which adopts estimated point-cloud-indexed retrieval following Lyra 2.0~\cite{shen2026lyra};
and (5) \textit{I3DM-based Retrieval}, which adopts implicit 3D-aware retrieval following I3DM~\cite{li2026i3dm}.
All three retrieval replace our Tile-indexed Memory and use maximum-coverage selection. 

As shown in Tab.~\ref{tab:ablation} (top), our full method outperforms all variants in multi-view consistency and camera control accuracy. 
As illustrated in Fig.~\ref{fig:ab_retrieval}, without memory, the model fails to preserve previously observed scene content. Removing spatially masked memory attention allows irrelevant historical content to interfere with generation (e.g., introducing kitchen-like content from the initial observation). FoV-based, geometry-based and I3DM-based retrieval struggle to retrieve relevant observations under large viewpoint changes, leading to hallucinated content and scene inconsistencies. In contrast, our full method effectively retrieves and utilizes relevant historical content, maintaining scene consistency even under complex exploration trajectories. We further evaluate the runtime efficiency of Geo-based, I3DM-based, and our Tile-indexed Memory on the same hardware. Specifically, we generate 972-frame videos across five scenes, each consisting of four clips, and measure the average memory writing and retrieval time per clip. Geo-based Memory takes 9.289 s/clip (writing/retrieval: 7.878/1.411 s), while I3DM-based Memory takes 5.54 s/clip (writing/retrieval: 0.0003/5.536 s). In contrast, our memory requires only 0.088 s/clip (writing/retrieval: 0.036/0.052 s), demonstrating the efficiency of floorplan-based spatial indexing.

\paragraph{Layout-aware Guidance.}
We evaluate our Layout-aware Guidance in Tab.~\ref{tab:ablation} (bottom) and Fig.~\ref{fig:ab_guidance} using two variants:
(1) \textit{w/o Layout Guidance}, which replaces our Layout-aware Guidance with lattice-rendering guidance following Wonder~\cite{xu2026wonder};
(2) \textit{w/o Semantic Labels}, which removes semantic labels, retaining only coarse geometric guidance. As shown in Fig.~\ref{fig:ab_guidance} (top), using only camera-motion guidance without explicit architectural constraints leads to inaccurate camera navigation and structural collapse. In contrast, our Layout-aware Guidance produces more plausible results with consistent point-cloud reconstructions.
In Fig.~\ref{fig:ab_guidance} (bottom), removing semantic labels causes abrupt scene changes during room transitions. For example, the video model hallucinates a door abruptly on the generated scene when entering new rooms, causing incoherent video generation.
Furthermore, semantic cues can help the model localize and attend to relevant historical content in the retrieved memory, leading to improved scene consistency, as shown in Tab.~\ref{tab:ablation} (bottom).

\section{Conclusion}


In this work, we presented \textbf{IndoorMem}, a floorplan-based framework for consistent long-horizon indoor video generation. Our key insight is that a single coarse 2D floorplan can serve as both a spatial index for recalling the past and a structural blueprint for guiding the future. Building on this shared scaffold, Tile-indexed Memory enables efficient, room-level occlusion-aware retrieval; Layout-aware Guidance provides geometric and semantic cues for camera control; and Spatially Masked Memory Attention suppresses interference from irrelevant historical content. Extensive experiments demonstrate superior multi-view consistency and camera control accuracy while maintaining high video quality, highlighting the potential of lightweight spatial scaffolds for explorable indoor video worlds.

{
    \small
    \bibliographystyle{ieeenat_fullname}
    \bibliography{main}
}


\end{document}


\maketitlesupplementary
\raggedbottom

\noindent This supplementary material is organized as follows:
\begin{itemize}[leftmargin=1.25em,itemsep=2pt,topsep=4pt]
  \item Sec.~\ref{sec:suppl_limitations}: \textbf{Limitations and Discussion.}
  \item Sec.~\ref{sec:suppl_floorplan}: \textbf{Examples of Floorplan Initialization.}
  \item Sec.~\ref{app:tile_memory}: \textbf{Details of Tile-indexed Memory.}
  \item Sec.~\ref{sec:suppl_implementation}: \textbf{Implementation Details.}
  \item Sec.~\ref{sec:suppl_comparisons}: \textbf{More Comparisons.}
  \item Sec.~\ref{sec:suppl_results}: \textbf{More Results and Applications.}
\end{itemize}

\section{Limitations and Discussion}
\label{sec:suppl_limitations}
Despite the effectiveness of our method, several limitations remain.

\begin{figure}
  \centering
  \includegraphics[width=\linewidth]{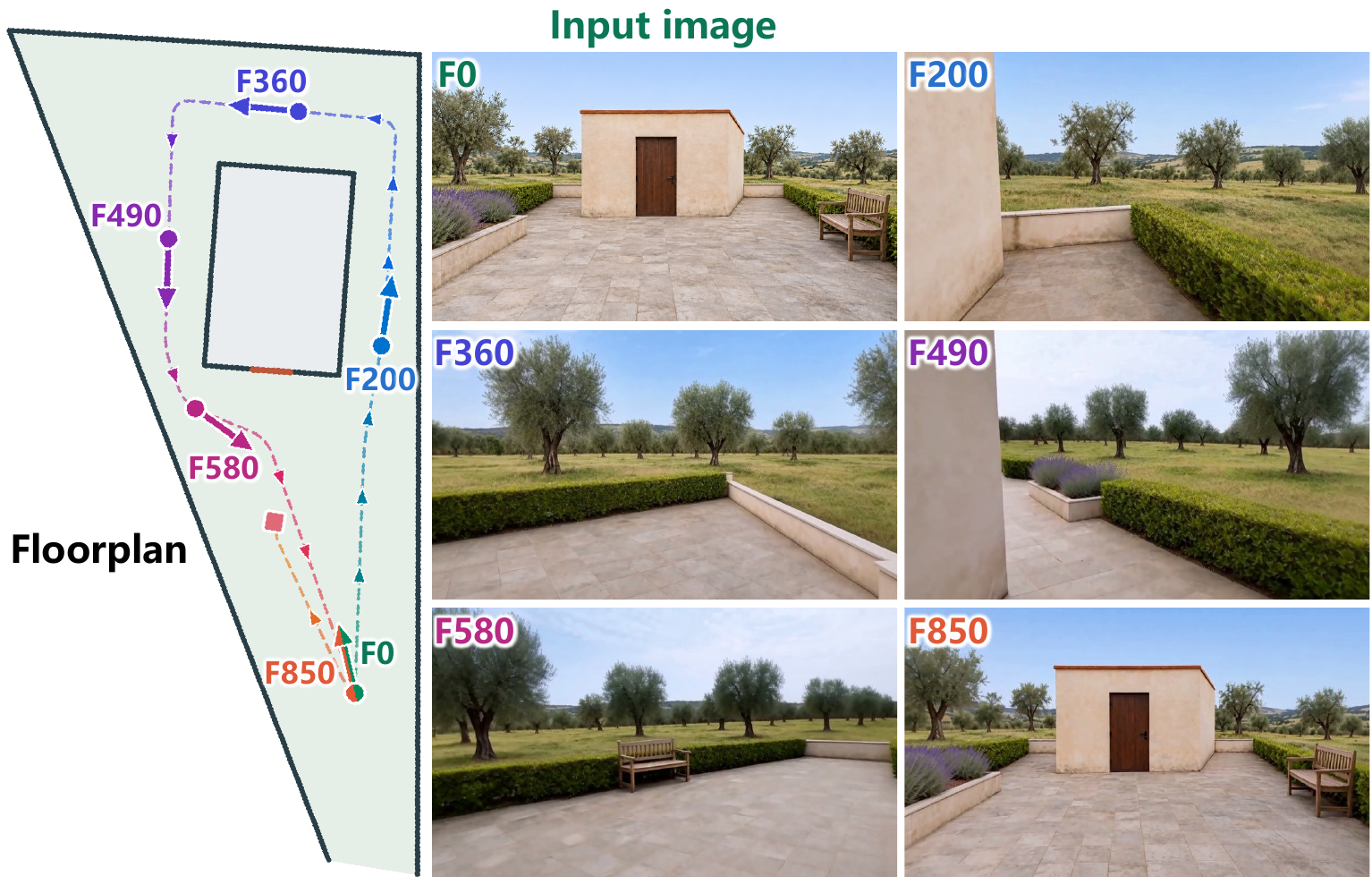}
  \caption{\textbf{Example of an outdoor scene.} Given an input image, we use an MLLM to generate a floorplan following our specification. Without any outdoor-specific fine-tuning, our method can generate an outdoor scene using this floorplan, suggesting its potential for extension to outdoor environments.}
  \label{fig:outdoor}
\end{figure}

\paragraph{Outdoor Environments.}
Our current framework primarily focuses on indoor video generation, but we also present a simple outdoor example in Fig.~\ref{fig:outdoor} without any outdoor-specific fine-tuning. During inference, we modify the text prompt to describe outdoor semantics, including interpreting ceiling-labeled regions as sky and wall-labeled regions as solid structures. This suggests that our method can generalize to outdoor scenes to some extent.
Although the underlying idea of using a coarse 2D layout as a spatial scaffold for memory and guidance could also be applied to outdoor video generation, e.g., using coarse bird's-eye-view (BEV) maps containing major landmarks and structural boundaries, extending it to more general outdoor environments would require more diverse structural and semantic representations. We leave this extension for future work.

\begin{figure}
  \centering
  \includegraphics[width=\linewidth]{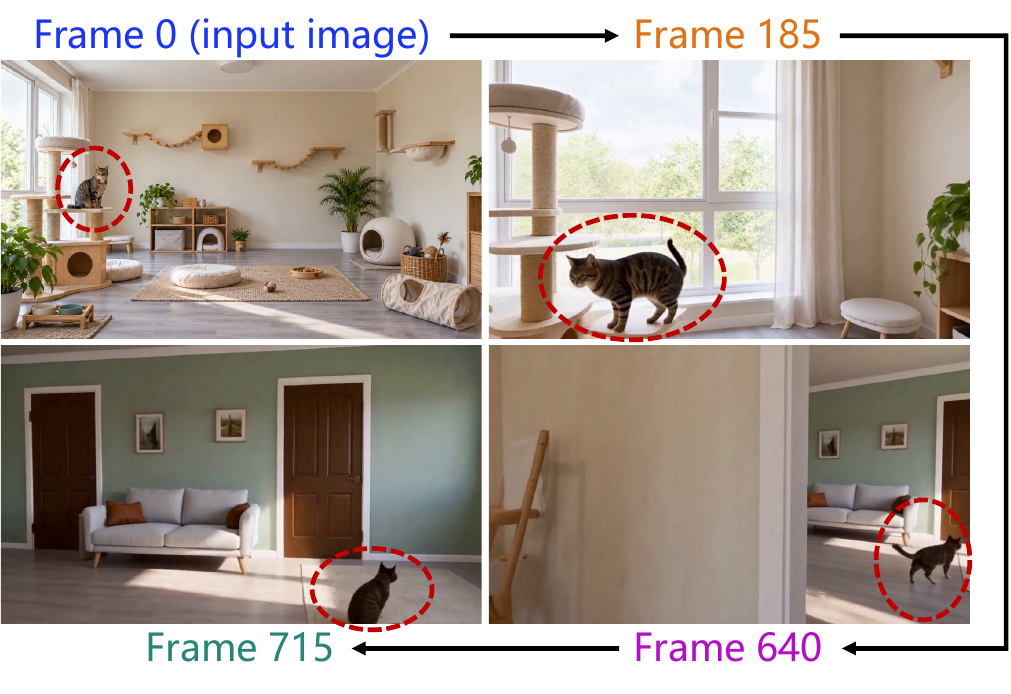}
  \caption{\textbf{Example containing a dynamic object.} Our method preserves the appearance of a dynamic object to some extent, suggesting its potential for handling dynamic scenes.
  }
  \label{fig:dynamic}
\end{figure}

\paragraph{Dynamic Scenes.}
Our current method mainly focuses on static scenes and supports only simple dynamic interactions, e.g., door-opening motion. However, since we do not explicitly restrict the model to generating static objects, dynamic objects may still appear in the generated videos. Interestingly, we observe that our method can preserve the appearance of dynamic objects to some extent, e.g., the cat in Fig.~\ref{fig:dynamic}. Extending our method to more general dynamic scenes could involve tracking and modeling dynamic objects and their motion within the floorplan, together with a dynamic memory bank to preserve their appearance over time.

\paragraph{Generation Time.}
Although our memory mechanism is efficient, generating a 243-frame video clip at $832\times480$ still takes approximately 3 minutes on eight H200 GPUs due to the VAE encoding/decoding and diffusion process. Further acceleration may require more efficient architectures, e.g., linear attention~\cite{chen2026sana}, optimized inference pipelines~\cite{li2026sol}, or distillation strategies~\cite{huang2026self, liu2026rolling}. We leave these extensions for future work.

\section{Examples of Floorplan Initialization}
\label{sec:suppl_floorplan}
\begin{figure}[tbp]
  \centering
  \includegraphics[width=\linewidth]{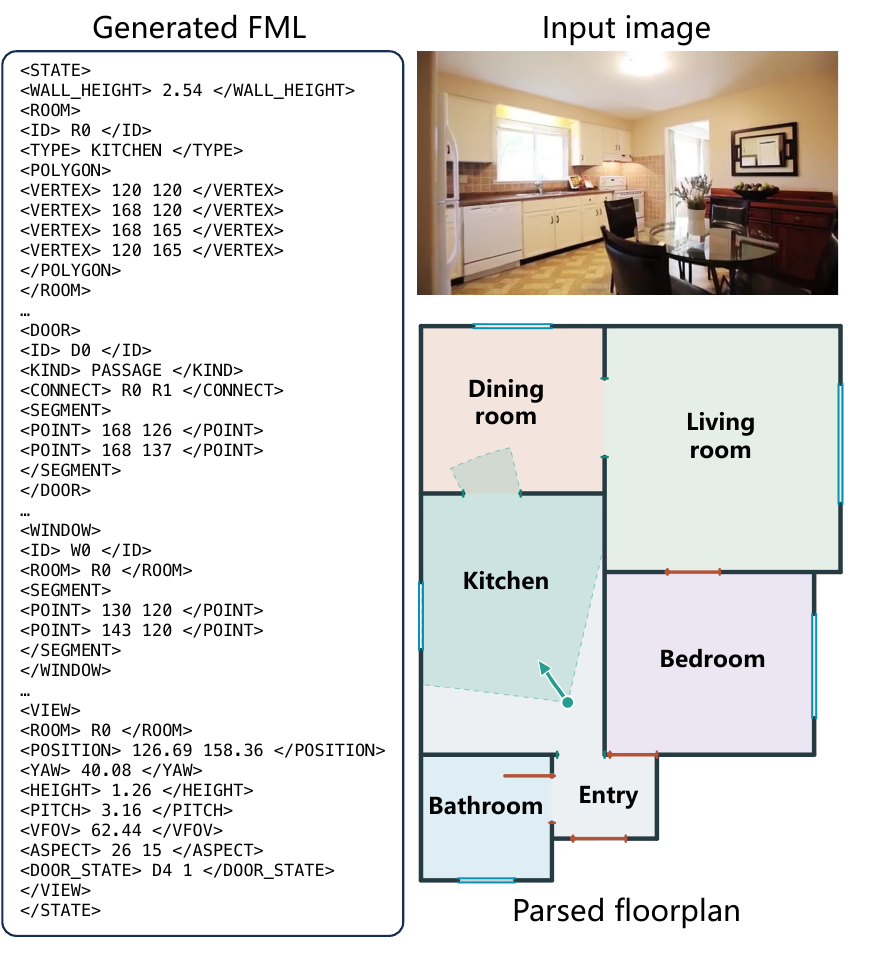}
  \caption{\textbf{Example of floorplan generation.} Given an initial input image, the provided specification and prompts, the MLLM generates an FML sequence that can be parsed into a 2D floorplan.
  }
  \label{fig:fml}
\end{figure}

We provide the specification of our extended Floorplan Markup Language (FML)~\cite{Shiohara_2026_CVPR} and the prompt for generating floorplans from initial images following the FML grammar in the supplementary files \texttt{specification.md} and \texttt{prompt\_from\_image.md}, respectively. Using these files and a user-provided initial image, we can prompt MLLMs (tested on GPT-6~\cite{OpenAI2026ChatGPT} and Qwen3.8-27B~\cite{qwen38}) to generate an extended FML sequence, which is then parsed into a 2D floorplan, as shown in Fig.~\ref{fig:fml}.

\section{Details of Tile-indexed Memory}
\label{app:tile_memory}
We provide details of the observation weights and relevance scoring used for memory retrieval.

\paragraph{Observation Weights.}
For each visible tile, the observation weight combines exponential decay terms for angular offset and distance:
\begin{equation}
\label{eq:ob_w}
    \psi_i(c)
    =
    \exp\!\left(-\frac{a_i(c)}{\sigma_a}-\frac{d_i(c)}{\sigma_d}\right),
\end{equation}
where $a_i(c)$ and $d_i(c)$ denote the angular offset and camera-to-tile distance, respectively. The parameters $\sigma_a$ and $\sigma_d$ control the decay scales. The angular scale is set relative to the camera's horizontal field of view.

\paragraph{Relevance Scoring for Retrieval.}
After obtaining the set of target-visible tiles $\mathcal{T}_Q$ through visibility testing under the target trajectory $Q$, we select a representative target camera $t^\star(c)$ for each tile $c$:
\begin{equation}
\begin{aligned}
    t^\star(c)
    &=
    \arg\max_{t \in \mathcal{A}(c)} \bar{\psi}_t(c), \\
    \bar{\psi}_t(c)
    &=
    \frac{\psi_t(c)}
    {\sum_{c' \in \mathcal{T}_{\mathcal{Q}}} \psi_t(c') + \epsilon},
\end{aligned}
\end{equation}
where $\mathcal{A}(c)$ denotes the set of target cameras observing tile $c$, $\psi_t(c)$ denotes the observation weight of tile $c$ at target camera $t$ computed using Eq.~\ref{eq:ob_w}, and $\bar{\psi}_t(c)$ is the normalized observation weight.

We measure viewing compatibility between historical camera $i$ and the representative target camera $t^\star(c)$ observing the same tile $c$, considering their positional distance $\Delta d$ on the floorplan and angular difference $\Delta\theta$ between their camera-to-tile viewing directions. The bounded decay functions are defined as
\begin{equation}
\begin{aligned}
    w_{\mathrm{pos}}(\Delta d)
    &=
    \eta_d + (1-\eta_d)\exp(-\frac{\Delta d}{\sigma_d}), \\
    w_{\mathrm{angle}}(\Delta\theta)
    &=
    \eta_\theta + (1-\eta_\theta)
    \exp\!\left(-\frac{{\Delta\theta}^2}{2\sigma_\theta^2}\right),
\end{aligned}
\end{equation}
where $\sigma_d$ and $\sigma_\theta$ control the decay scales, and $\eta_d$ and $\eta_\theta$ provide positive lower bounds.

The relevance score of candidate historical frame $i$ for target-visible tile $c$ is defined as
\begin{equation}
\label{eq:app_relevance}
\begin{aligned}
    r_i(c\mid\mathcal{Q})
    &=\psi_i(c)\,\kappa_i(c\mid\mathcal{Q}),\\
    \kappa_i(c\mid\mathcal{Q})
    &=\sqrt{
    w_{\mathrm{pos}}(\Delta d_i(c))
    \cdot
    w_{\mathrm{angle}}(\Delta\theta_i(c))
    }.
\end{aligned}
\end{equation}
We use the relevance scores for subsequent greedy maximum-coverage selection.

\section{Implementation Details}
\label{sec:suppl_implementation}

\begin{figure}[tbp]
  \centering
  \includegraphics[width=\linewidth]{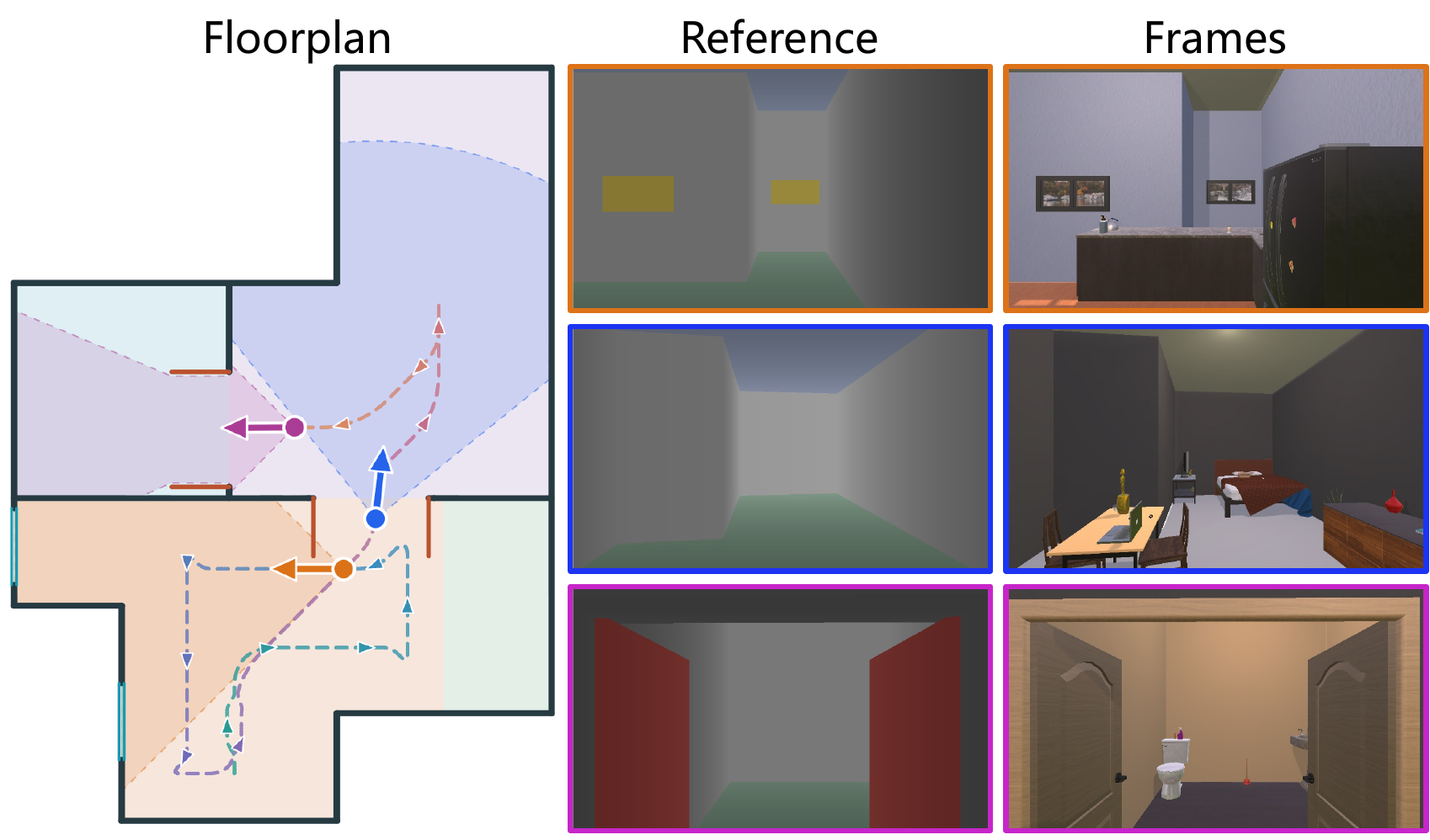}
  \caption{\textbf{Example of training data.}
  Left: 2D floorplan and camera trajectory; Middle: Rendered hybrid geometric and semantic guidance used as a reference video; Right: Rendered RGB video frames.
  }
  \label{fig:training}
\end{figure}

\begin{table*}[!t]
\centering
\caption{\textbf{Quantitative comparison}. Rev./Adj. MEt3R denote revisit/adjacent-frame MEt3R, and ZOR denotes zero-overlap rate. Overall averages SC, BC, TF, MS, and AQ.
Best and second-best results are shown in bold and underlined, respectively.}
\label{tab:quantitative}

\setlength{\tabcolsep}{2.5pt}
\renewcommand{\arraystretch}{1.15}

\footnotesize

\begin{tabular*}{\textwidth}{@{\extracolsep{\fill}}l*{12}{c}@{}}

\toprule
& \multicolumn{3}{c}{Multi-view Consistency}
& \multicolumn{3}{c}{Camera Control}
& \multicolumn{6}{c}{Video Generation Quality} \\
\cmidrule(lr){2-4}
\cmidrule(lr){5-7}
\cmidrule(l){8-13}

Method
& Rev.\ MEt3R$\downarrow$
& ZOR (\%)$\downarrow$
& Adj.\ MEt3R$\downarrow$
& RotErr$\downarrow$
& TransErr$\downarrow$
& CamMC$\downarrow$
& SC$\uparrow$
& BC$\uparrow$
& TF$\uparrow$
& MS$\uparrow$
& AQ$\uparrow$
& Overall$\uparrow$ \\
\midrule

WorldMesh
& 0.562 & \underline{8.54} & 0.063
& \underline{30.92} & \underline{1.211} & \underline{1.448}
& 71.03 & 82.29 & 93.66 & 98.31
& 47.75 & 78.61 \\

Code-World-Model
& \underline{0.533} & 21.41 & \underline{0.052}
& 96.58 & 2.480 & 3.291
& \textbf{78.97} & \textbf{88.17} & \underline{94.84} & \underline{98.33}
& \textbf{49.64} & \textbf{81.99} \\

\midrule
\textbf{Ours}
& \textbf{0.425} & \textbf{5.81} & \textbf{0.038}
& \textbf{28.58} & \textbf{1.207} & \textbf{1.410}
& \underline{78.61} & \underline{87.08} & \textbf{95.88}
& \textbf{99.02} & \underline{48.71} & \underline{81.86} \\

\bottomrule
\end{tabular*}
\end{table*}

\paragraph{Datasets.}
For training, we construct a synthetic indoor exploration dataset by rendering 675 procedurally modeled scenes from ProcTHOR~\cite{procthor}, obtaining 1,586 camera trajectories, each containing 972 frames with revisits. An example is shown in Fig.~\ref{fig:training}. We sample six 243-frame clips per trajectory using stratified random sampling, yielding 9,516 training clips in total. We generate text captions for these clips using Qwen3.8-27B~\cite{qwen38}.

For evaluation, we collect 407 initial indoor images, including 207 generated by GPT-Image 2 and 200 real images from RealEstate10K~\cite{Zhou2018} and DL3DV~\cite{ling2024dl3dv}, covering diverse indoor environments and visual styles. For each image, we employ GPT-6~\cite{OpenAI2026ChatGPT} to initialize a floorplan, based on which we construct a corresponding 972-frame camera trajectory with revisits, yielding 407 long-horizon evaluation sequences.

\paragraph{Implementation Details.}
For training, we fine-tune the MiniMax-H3 Ref2VA~\cite{MiniMaxH3_2026} backbone using rank-32 LoRA~\cite{hu2021lora} adapters applied to the attention and feed-forward layers of the DiT. We optimize only the LoRA parameters using AdamW with a learning
rate of $10^{-4}$, keeping the pretrained parameters frozen. Training runs for 1,000 steps on eight NVIDIA H200 GPUs with a total batch size of eight, taking approximately 1.5 days.
Each training sample contains 243 frames at a resolution of $832\times480$. All samples include text and layout guidance, while 10\% use no image references, 10\% additionally use the first image, and 80\% use both the first image and up to eight historical frames retrieved by our Tile-indexed Memory (i.e., $K=8$).

At inference, we generate long-horizon videos autoregressively in 243-frame clips, using the final frame of each clip as the first image condition for the next. From the second clip onward, we retrieve up to eight historical memory frames. We use 50 denoising steps per clip, with LoRA scales of 0.25 for the first clip and 0.5 for subsequent clips.

\enlargethispage{-8\baselineskip}
\section{More Comparisons}
\label{sec:suppl_comparisons}

\begin{figure*}[p]
  \centering
  \includegraphics[width=\textwidth]{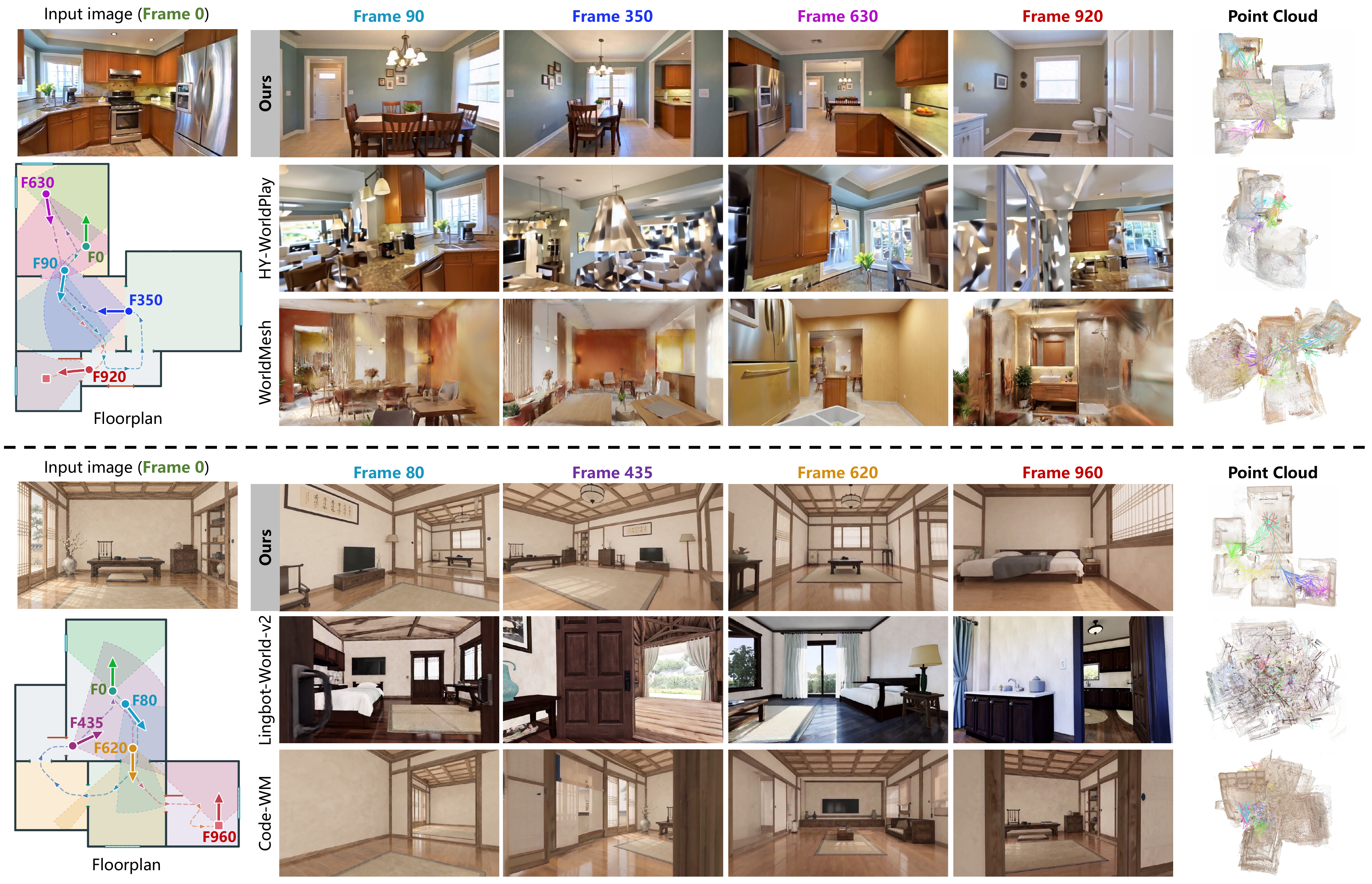}
  \caption{\textbf{Qualitative comparisons with other methods.} We compare generated frames and reconstructed point clouds across two indoor scenes. The input images and floorplans with camera trajectories are shown on the left. Our method better preserves scene appearance and structure during long-horizon exploration and revisits.
  }
  \label{fig:comp_1}
\end{figure*}

\begin{figure*}[p]
  \centering
  \includegraphics[width=\textwidth]{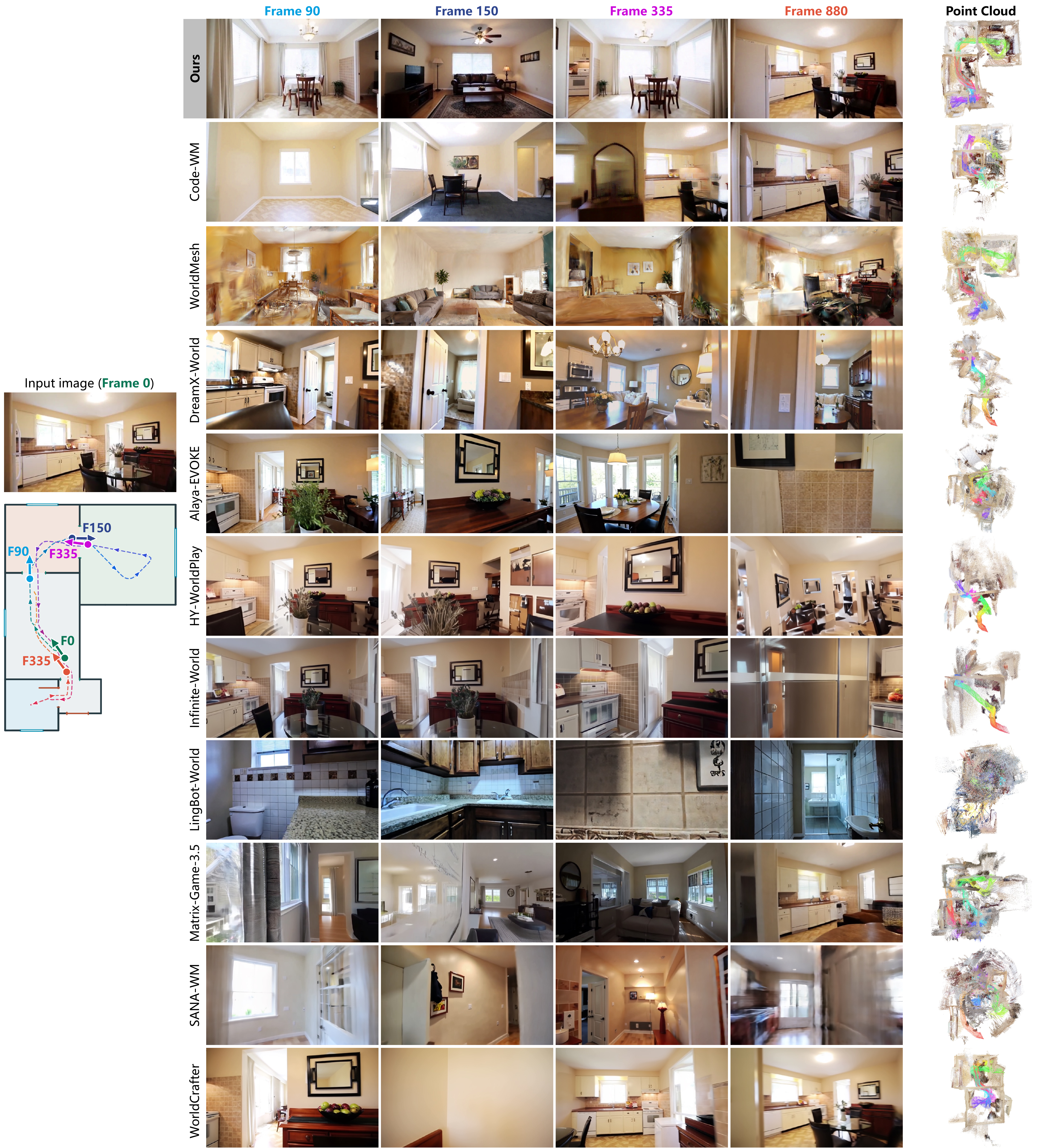}
  \caption{\textbf{Qualitative comparison with all baselines.} We compare generated frames and reconstructed point clouds with 11 baseline methods. The input images and floorplans with camera trajectories are shown on the left. Our method better preserves scene appearance and structure during long-horizon exploration and revisits.}
  \label{fig:comp_2}
\end{figure*}

We additionally compare with Code-World-Model~\cite{chen2026code}, a recent method that uses a coding agent to generate proxy videos as references for video generation, and WorldMesh~\cite{schneider2026worldmesh}, which uses coarse meshes to guide multi-room 3D scene generation with 3DGS~\cite{kerbl3Dgaussians}. The quantitative results are shown in Tab.~\ref{tab:quantitative}, while qualitative comparisons are presented in Fig.~\ref{fig:comp_1} and Fig.~\ref{fig:comp_2}.
Our method outperforms both baselines in multi-view consistency and camera control accuracy while achieving comparable overall video quality to Code-World-Model. As shown in the qualitative comparisons (Fig.~\ref{fig:comp_1} and Fig.~\ref{fig:comp_2}), Code-World-Model can only maintain consistency with the first input frame but fails to preserve subsequently generated content. WorldMesh suffers from severe rendering artifacts, particularly in insufficiently covered regions. In contrast, our method maintains scene consistency while preserving high generation quality.

\section{More Results and Applications}
\label{sec:suppl_results}

\begin{figure*}[p]
  \centering
  \includegraphics[width=\textwidth]{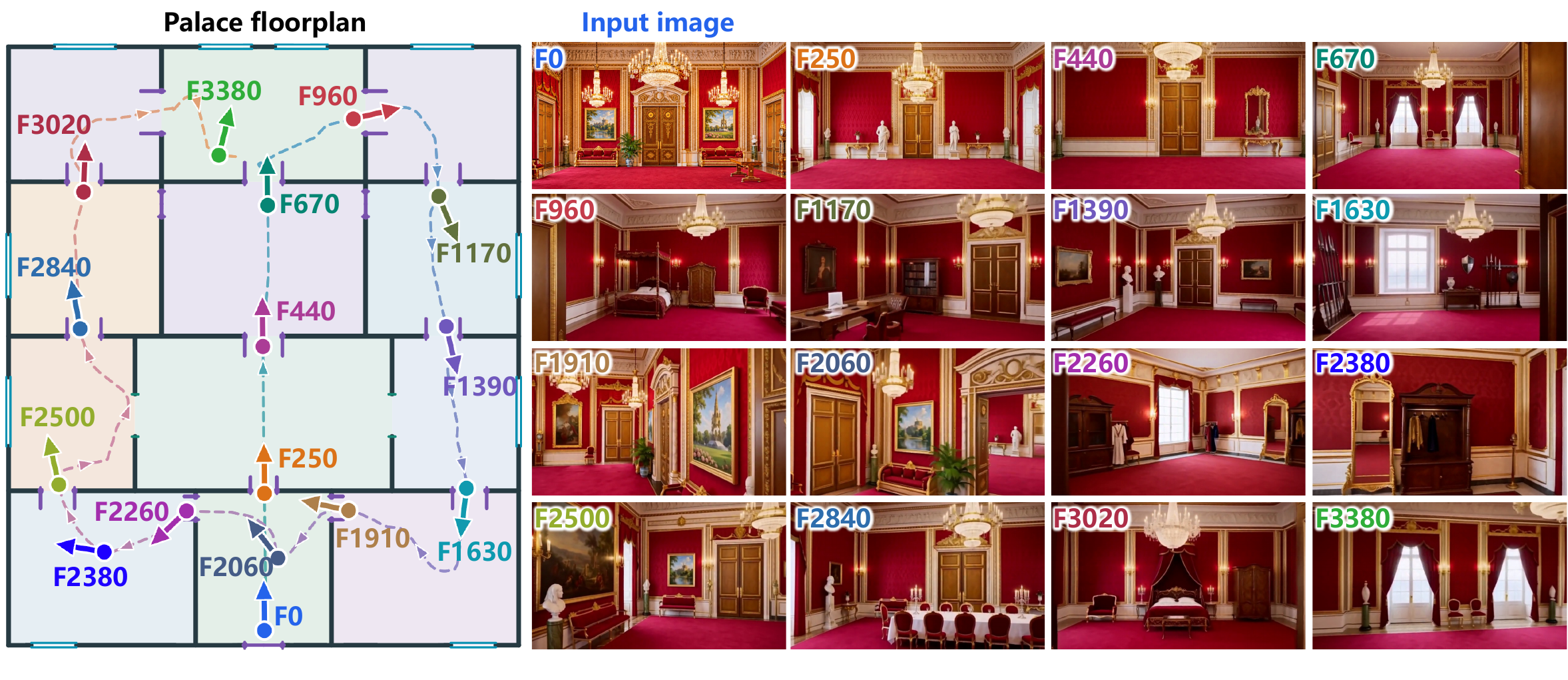}
  \caption{\textbf{Large-scale scene generation.} Given a single input image, our method generates a large-scale indoor scene--\textbf{a palace}--and supports exploration over 3,380 frames (approximately 2.5 minutes) while maintaining consistency when revisiting previously generated regions.
  }
  \label{fig:long}
\end{figure*}

\paragraph{Large-scale Scene Generation.} As shown in Fig.~\ref{fig:long}, we demonstrate large-scale indoor scene generation in a European palace comprising 12 rooms. Starting from a single input image, we generate 3,388 frames (approximately 2.5 minutes) in total. Our method maintains realistic appearance and scene consistency throughout this long sequence, demonstrating its potential for game level generation.

\begin{figure*}[p]
  \centering
  \includegraphics[width=0.95\textwidth]{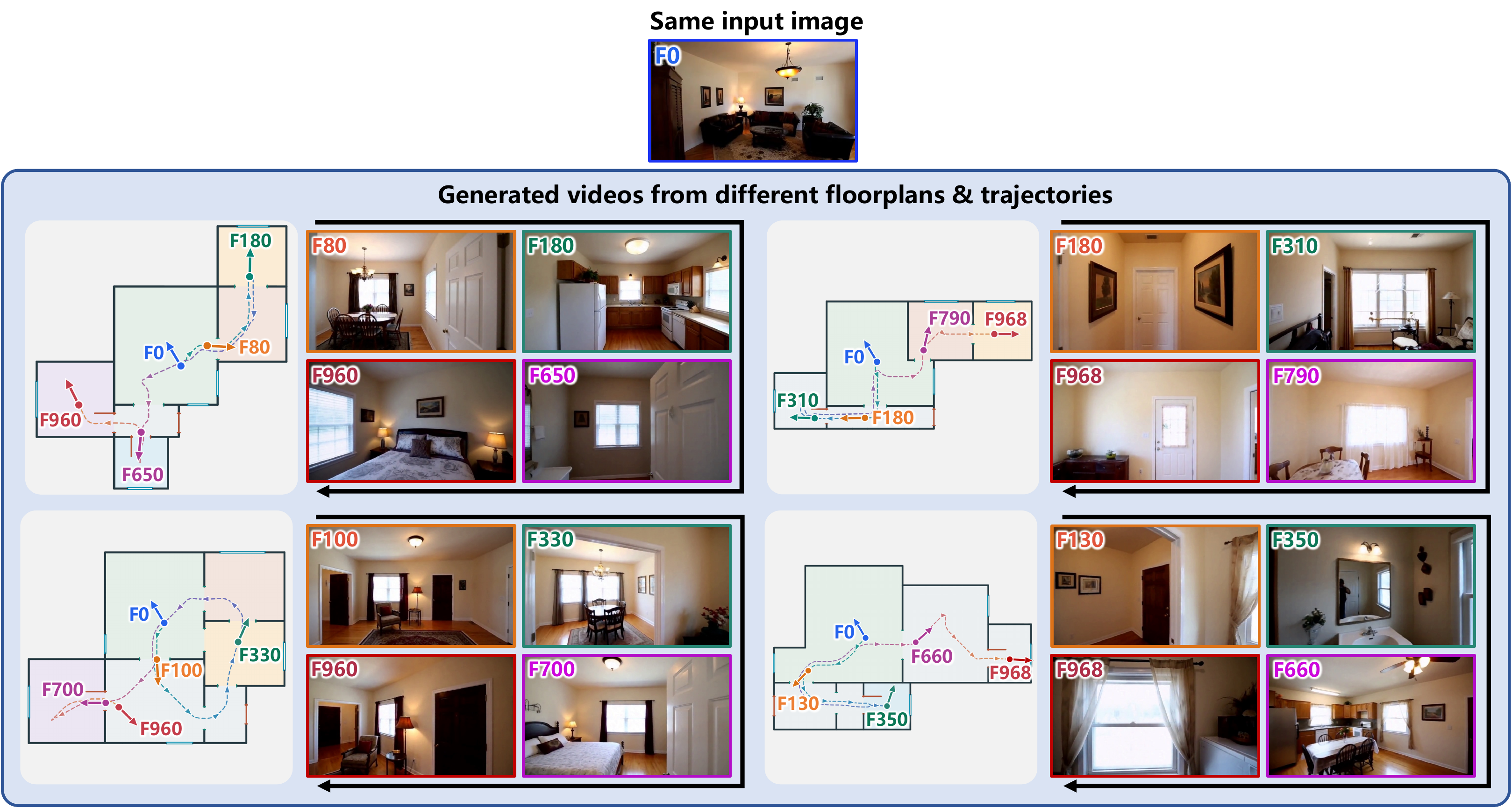}
  \caption{\textbf{One image, multiple floorplans.} Given the same input image but different floorplans, our method generates different indoor scenes and exploration results.
  }
  \label{fig:variety_1}
\end{figure*}
\begin{figure*}[p]
  \centering
  \includegraphics[width=0.95\textwidth]{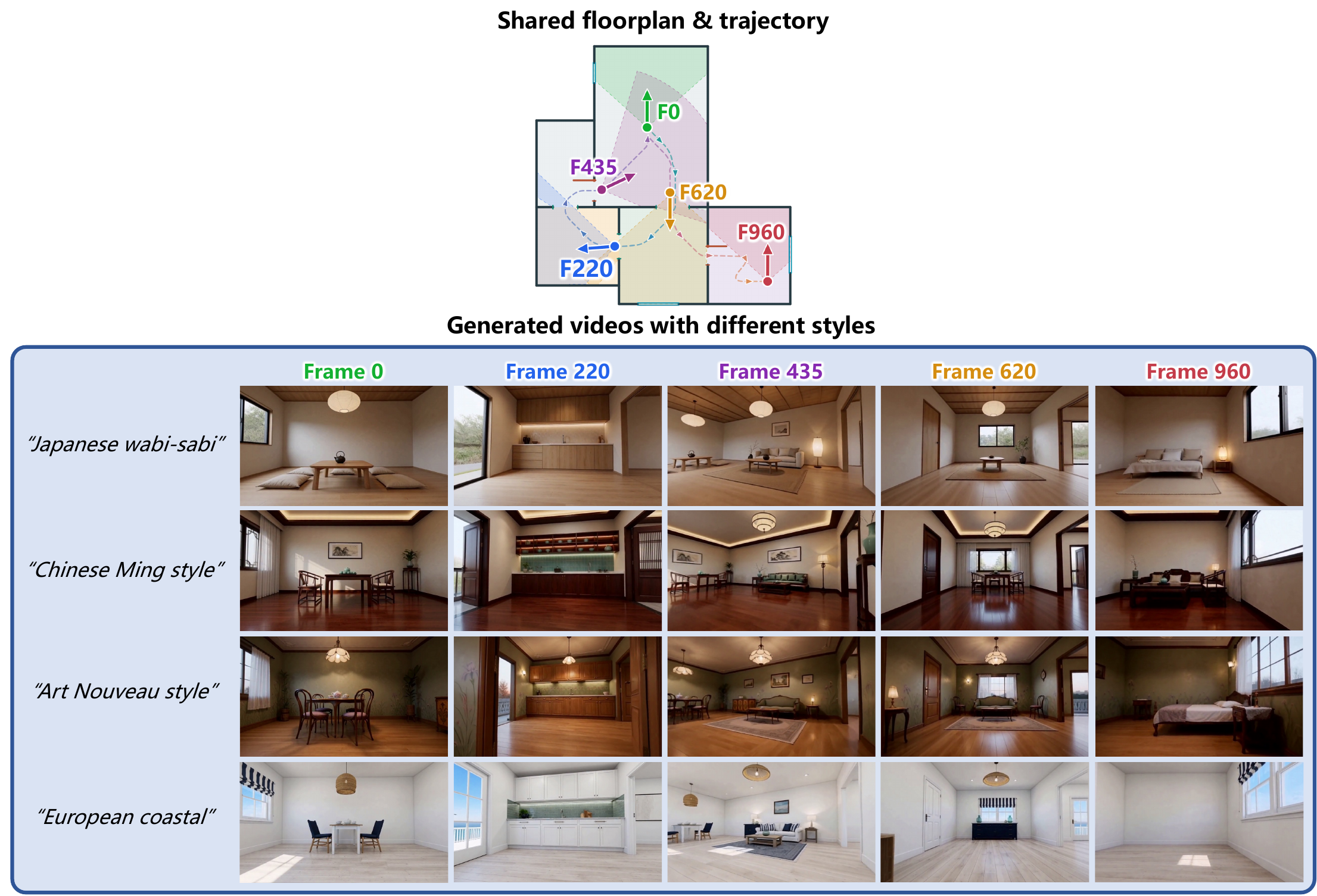}
  \caption{\textbf{One floorplan, multiple styles.} Given the same floorplan and camera trajectory, our method supports text-to-video generation. With different styles specified in the text prompts, our method generates indoor scenes with substantially different appearances.
  }
  \label{fig:variety_2}
\end{figure*}

\paragraph{Generative Diversity.}
Beyond long-horizon consistency, our method also supports diverse indoor world generation through flexible control of spatial layouts and visual appearances. As shown in Fig.~\ref{fig:variety_1}, starting from the same input image, we generate four distinct indoor scenes using different floorplans and camera trajectories, demonstrating the flexibility of exploring diverse spatial layouts from a single observation. Furthermore, given the same floorplan and camera trajectory, our method supports text-to-video generation with different visual styles specified by text prompts, as shown in Fig.~\ref{fig:variety_2}. This enables diverse scene appearances while maintaining the same underlying architectural layout. These results highlight the potential of our method for flexible indoor environment design.

\clearpage

\clearpage
\makeatletter
\global\@colht\textheight
\global\@colroom\textheight
\global\vsize\textheight
\makeatother
\enlargethispage{-24\baselineskip}
{
    \small
    \bibliographystyle{ieeenat_fullname}
    \bibliography{main}
}
